\pdfoutput=1

\documentclass[11pt]{article}

\usepackage[final]{acl}

\usepackage{times}
\usepackage{latexsym}
\usepackage[T1]{fontenc}
\usepackage[utf8]{inputenc}
\usepackage[protrusion=true,expansion=alltext,stretch=20,shrink=20]{microtype}
\usepackage{inconsolata}

\usepackage{graphicx}
\usepackage{hyperref}
\usepackage{url}
\usepackage{booktabs}
\usepackage{subcaption}
\usepackage{amsmath}
\usepackage{amssymb}
\usepackage{multirow}
\usepackage{makecell}
\usepackage{colortbl}
\usepackage{xcolor}
\usepackage{wrapfig}
\usepackage{placeins}
\usepackage{tcolorbox}
\usepackage{xspace}

\renewcommand{\topfraction}{0.85}
\renewcommand{\bottomfraction}{0.7}
\renewcommand{\textfraction}{0.15}
\renewcommand{\floatpagefraction}{0.7}
\renewcommand{\dbltopfraction}{0.85}
\renewcommand{\dblfloatpagefraction}{0.7}
\title{\name: Unmasking the Impact of Transliteration on Bangla Dialectal LLMs}

\author{
\textbf{Md Mahir Jawad\textsuperscript{1, 2}},
\textbf{Galib Mahmud Jim\textsuperscript{2}},
\textbf{Rafid Ahmed\textsuperscript{3}}, \\
\textbf{Mir Sazzat Hossain\textsuperscript{2, 4}},
\textbf{Md Fahim\textsuperscript{2}},
\textbf{Md Farhad Alam Bhuiyan\textsuperscript{2}}\\
\\
\textsuperscript{1}\textit{BRAC University},\\
\textsuperscript{2}\textit{Penta Global Limited},\\
\textsuperscript{3}\textit{University of Central Florida},\\
\textsuperscript{4}\textit{Center for Computational \& Data Sciences, Independent University, Bangladesh}\\
}

\newcommand{\name}{\textsc{5-Dialects-BN}\xspace}

\begin{document}
\maketitle
\begin{abstract}
Large Language Models (LLMs) have achieved remarkable progress across natural language processing (NLP) tasks, yet their capabilities degrade sharply for low-resource languages and dialectally diverse settings. Bangla, the world's sixth most spoken language, exemplifies this gap: existing resources overwhelmingly target Standard Bangla, leaving its regional dialects without the benchmarks needed to develop or evaluate dialect-aware systems.
We address this gap with \name, the first multi-annotation Bangla dialect benchmark to align Romanized transliteration with dialectal text, Standard Bangla, English, and subjectivity labels across five regional varieties. The dataset comprises 6{,}000 manually annotated entries spanning five major dialects: Chittagong, Barisal, Noakhali, Sylhet, and Rangpur (Chittagong 1{,}900; Noakhali 1{,}500; Sylhet 1{,}200; Barisal 700; Rangpur 700), reflecting natural online availability. Each entry is enriched with five aligned annotations: the original dialectal text, a Romanized transliteration, an English translation, a Standard Bangla translation, and a subjectivity label (subjective vs.\ objective). Annotations were produced and cross-validated by native speakers and undergraduate linguistics students to ensure dialectal authenticity and semantic fidelity.
The resulting resource supports a diverse suite of tasks, including dialect identification, dialect-to-standard normalization, machine translation, subjectivity classification, and parameter-efficient fine-tuning (\emph{e.g.}, LoRA) of multilingual LLMs. By providing a standardized, multi-annotation benchmark, \name enables principled evaluation of LLMs on dialectally diverse Bangla and lays a foundation for further research in low-resource, dialect-aware NLP.
\end{abstract}

\section{Introduction}

Large Language Models (LLMs) have driven rapid progress across natural language processing (NLP), but this progress is unevenly distributed across the world's languages: low-resource and dialectally diverse linguistic communities remain underserved by both training corpora and evaluation benchmarks~\cite{joshi2020state,blasi2022systematic}.
Bangla, spoken by over 270 million people, exemplifies this gap~\cite{bhattacharjee2022banglabert}.
While Standard (Cholito) Bangla has received growing attention in diverse NLP tasks~\cite{ahmed-etal-2026-evaluating, tinyllm2025dehan}, its regional dialects (which differ markedly from the standard in phonology, lexicon, morphology, and syntax~\cite{grierson1903linguistic,mahjabin2025human}) remain critically under-resourced, and the absence of multi-annotation benchmarks makes it difficult to diagnose model failures or evaluate dialect-aware systems in a reproducible manner.

\begin{figure*}[!t]
\centering
\includegraphics[width=0.8\textwidth]{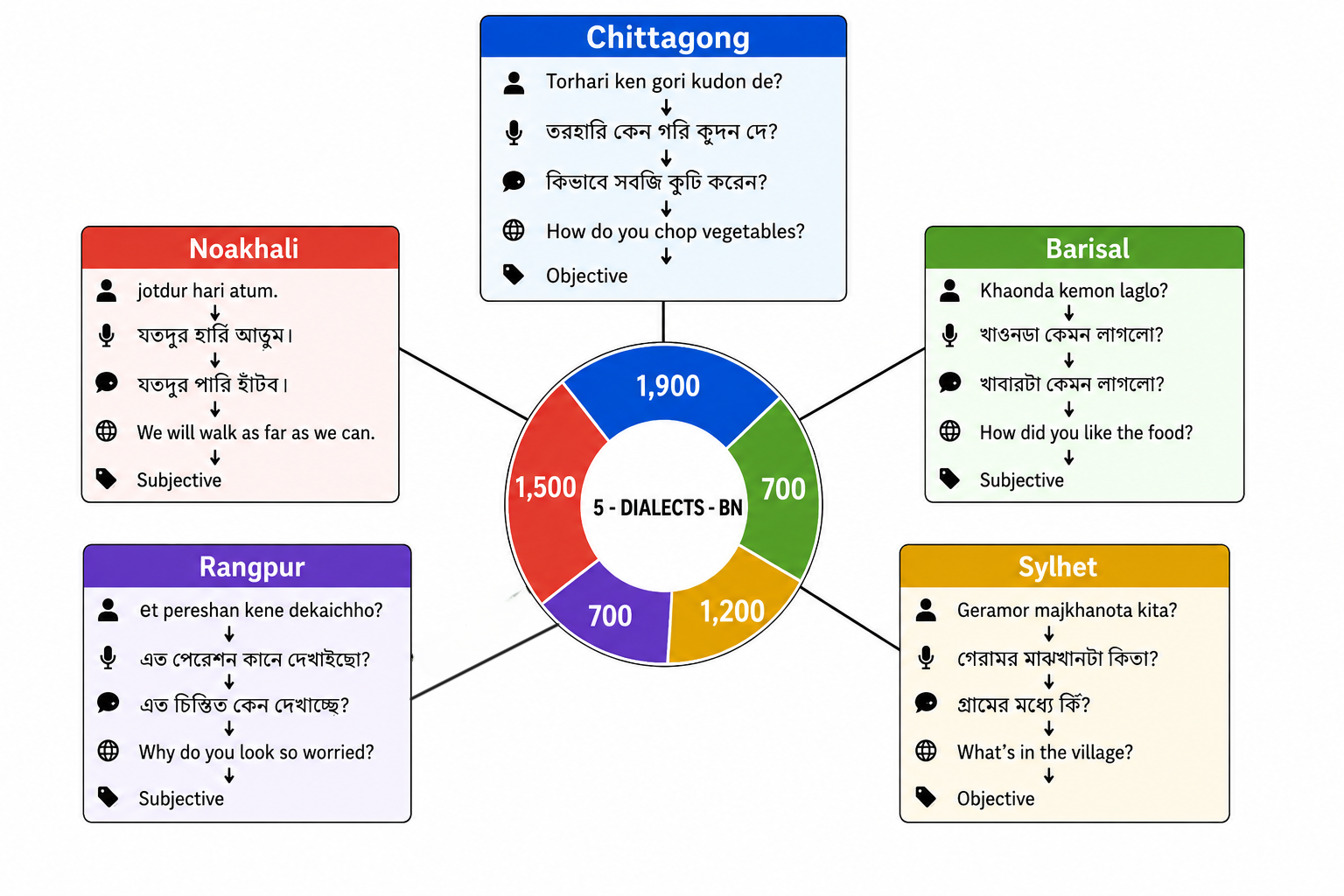}
\caption{Representative entries from each of the five regional dialects in \name. Each entry supplies five aligned fields: dialectal text in native Bangla script, a Romanized transliteration, a transliteration normalization, a binary subjectivity label, and an English translation. The selected examples span both subjectivity classes and illustrate the lexical and phonological divergence each dialect introduces over Standard Bangla.}
\label{fig:dialect_examples}
\end{figure*}

To address this challenge, we introduce \name, a manually curated and verified benchmark containing $6{,}000$ utterances across five major regional Bangla dialects: Chittagong, Barisal, Noakhali, Sylhet, and Rangpur (Figure~\ref{fig:dialect_examples}).
Each entry provides five aligned fields: native dialectal text, Romanized transliteration, Standard Bangla translation, English translation, and a binary subjectivity label.
We evaluate seven contemporary LLMs across three core tasks: \textbf{(i)~dialect-to-English machine translation}, \textbf{(ii)~binary subjectivity classification}, and \textbf{(iii)~dialect-to-Standard-Bangla normalization}.
We benchmark these models across four evaluation regimes: zero-shot, few-shot, and chain-of-thought (CoT) prompting, alongside parameter-efficient fine-tuning via LoRA~\cite{hu2022lora} on open-source LLMs. This evaluation yields two primary findings:
\begin{enumerate}
    \item \textbf{Supervision bridges the resource divide:} Fine-tuning open-source models with LoRA on just $160$ examples per dialect surpasses every closed-source zero-shot and few-shot prompting baseline on translation and subjectivity classification (e.g., Mistral-7B reaching $73.6$ BLEU vs.\ Gemini~3~Flash's $46.4$ zero-shot).
    \item \textbf{Transliteration harms current LLMs:} In contrast to earlier reports on smaller multilingual models~\cite{khanuja2020gluecos,ma2024translit}, Romanized transliterated input consistently and severely degrades current LLM performance across all models and regimes, driven by many-to-one phonemic information loss and subword token fragmentation.
\end{enumerate}

\paragraph{Why transliteration is a first-class field.} Online users in South Asia overwhelmingly type dialectal Bangla in Latin script (Banglish) on platforms such as YouTube, Facebook, and Reddit~\cite{bali2014borrowing,fahim2024banglatlit,haider2025banth}.
Evaluating models on Romanized text is therefore essential for practical deployment.
By aligning Romanized transliteration as a first-class factorial variable alongside native script, \name{} enables the first causal decomposition of the transliteration penalty on modern LLMs (Section~\ref{sec:results}), revealing how orthographic ambiguity destroys underlying dialectal distinctions.

Our primary contributions are:
\begin{itemize}
    \item We release \name, the \textit{first multi-annotation Bangla dialect benchmark} to incorporate Romanized transliteration as an aligned field alongside dialectal text, Standard Bangla, English, and subjectivity labels across five regional dialects ($6{,}000$ verified entries).
    \item We establish inter-annotator reliability with Cohen's $\kappa = 0.78$ for subjectivity, $\kappa = 0.74$ for Standard Bangla rendering, and $0.84$ ROUGE-L agreement on English translations.
    \item We benchmark seven LLMs across zero-shot, few-shot, CoT prompting, and LoRA fine-tuning under native and Romanized input across three tasks, providing an empirical and causal analysis of script effects, prompting failure modes (demonstration interference and format collapse), and linguistic divergence.
\end{itemize}
\section{Related Work}
\label{sec:related}

\paragraph{Dialectal Bangla resources.} Recent efforts have introduced parallel and annotated resources for regional Bangla dialects, including Vashantor~\cite{faria2023vashantor}, ONUBAD~\cite{sultana2025onubad}, ChatgaiyyaAlap~\cite{chowdhury2025chatgaiyyaalap}, and ANCHOLIK-NER~\cite{paul2025ancholik} covering subsets of Chittagong, Sylhet, and Barisal. Closest to our work is DIALTSA-BN~\cite{jawad2025dialtsabn}, which compiles $600$ utterances across four dialects with Standard Bangla translations and sentiment labels. \name{} advances this line of work along multiple dimensions: (i)~expanding the corpus ten-fold to $6{,}000$ utterances; (ii)~incorporating the northwestern Rangpur dialect alongside Chittagong, Barisal, Noakhali, and Sylhet; (iii)~providing five fully aligned fields per utterance (adding Romanized transliteration and English translations); (iv)~evaluating script effects factorially across prompting and LoRA fine-tuning; and (v)~conducting causal tokenizer and linguistic error analyses (see Table~\ref{tab:app_dialtsa_comp} in Appendix~\ref{app:dialect_background} for a side-by-side comparison).

\paragraph{Bangla MT and dialect normalization.} Machine translation for Standard Bangla has progressed through large parallel benchmarks like BanglaNMT~\cite{hasan2020banglanmt} and Samanantar~\cite{ramesh2022samanantar}, alongside sequence-to-sequence architectures such as BanglaT5~\cite{bhattacharjee2023banglat5}, IndicBART~\cite{dabre2022indicbart}, and IndicTrans2~\cite{gala2023indictrans2}. However, these models are trained almost exclusively on formal Standard Bangla and fail to generalize to colloquial regional dialects. Our dialect-to-standard normalization benchmark parallels similar regional-to-standard mapping tasks in Arabic dialects (\citealp{zbib2012machine}) and Swiss German (\citealp{samardzic2016arcmad}).

\paragraph{Transliteration and script effects in Bangla.} BanglaTLit~\cite{fahim2024banglatlit} investigated back-transliteration and showed that transliteration-adapted encoders improve noisy social media text classification, while BanTH~\cite{haider2025banth} explored hate-speech detection in Romanized Bangla. \citet{improving2024ahmed} showed that fine-tuned transformers barely outperform classical TF-IDF baselines on transliterated Bangla, Hindi, and Arabic. While \citet{ma2024translit} observed in-context learning benefits from transliteration in older multilingual models, our study discovers an opposite trend on modern LLMs, which we systematically decompose in Section~\ref{sec:results}.

\paragraph{Affective text and evaluation targets.} SentNoB~\cite{islam2021sentnob} demonstrated that standard Bangla sentiment models degrade severely on noisy user-generated content. In \name, we evaluate dialect-to-English translation to leverage English as an orthographically stable scoring target~\cite{fahim2024banglatlit}, and use character-level (chrF++) and neural (COMET) metrics to evaluate dialect-to-Standard Bangla normalization without surface orthographic bias.

\section{\name}
\label{sec:dataset}

We construct \name{} through a four-stage curation and annotation pipeline: (i)~data harvesting from public online sources, (ii)~authenticity and boundary verification by native speakers, (iii)~tool-assisted multi-annotation by trained linguistics annotators, and (iv)~inter-annotator agreement validation.
Figure~\ref{fig:dialect_examples} previews representative entries across the five regional dialects, illustrating the five aligned fields per entry.

\paragraph{Data collection and dialect boundary verification.} Dialectal utterances are harvested from public online platforms: YouTube comment threads on regional media channels, regional community Facebook groups, Reddit discussions, and regional news features (Table~\ref{tab:app_sources}).
To guarantee clean dialect boundaries and single-label integrity, candidate utterances were screened by native speakers of each target dialect under Guideline G1 (Appendix~\ref{app:guidelines}): an utterance is accepted only if it contains $\ge 1$ dialect-diagnostic feature (lexical, morphological, or phonological), while ambiguous items or border cases along regional continua (e.g., Chittagong--Noakhali transitions) were flagged for panel adjudication.
Native-speaker verification rejected $34\%$ of harvested candidates on average.
Heavily code-mixed items were filtered out, while naturally occurring lexical borrowings were retained ($1$--$5\%$ of utterances contain $\ge 1$ Bangla-script English loanword).
Identifying spans (usernames, phone numbers, personal names) were manually stripped by verifiers; utterances remaining identifying after redaction were discarded.

\paragraph{Annotation tool and quality control.} Production annotation was conducted via a custom web-based tool (Figure~\ref{fig:annotation_ui}; live at \url{https://bangla-dialect-annotator.vercel.app/}) by fifteen undergraduate linguistics annotators (three native/proficient speakers per dialect), compensated above standard research-assistant rates.
For each row, the annotator inputs or verifies the native-script rendering via an integrated Avro input editor, verifies the Standard Bangla translation, edits or approves a draft English translation, and assigns a binary subjectivity label.
To mitigate annotator fatigue, candidate English translations were drafted by Gemini~\cite{comanici2025gemini} conditioned strictly on the human-verified Standard Bangla (not the noisy transliterated source).
Annotators actively edited $41\%$ of model drafts (ranging from $31\%$ in Rangpur to $52\%$ in Sylhet).
An independent control sample of $100$ items translated from scratch without AI drafts confirmed that pre-population introduced no stylistic anchoring bias (Appendix~\ref{app:iaa_details}).

\paragraph{Dataset composition and statistics.} Each entry contains five aligned fields: (1)~dialectal text in native Bangla script, (2)~Romanized transliteration, (3)~Standard Bangla translation, (4)~English translation, and (5)~a binary subjectivity label.
The dataset comprises $6{,}000$ verified entries spanning Chittagong ($1{,}900$), Noakhali ($1{,}500$), Sylhet ($1{,}200$), Barisal ($700$), and Rangpur ($700$), naturally reflecting online availability (Table~\ref{tab:app_dialect_dist}).
Utterance length ranges from $3$ to $40$ tokens (mean $11.2$, median $9$; Table~\ref{tab:utterance_stats}), capturing natural conversational dialogue.

\begin{table}[htbp]
\centering
\small
\begin{tabular}{lr}
\toprule
\textbf{Statistic} & \textbf{Value (Tokens)} \\
\midrule
Minimum Length & 3 \\
Maximum Length & 40 \\
Mean Length & 11.2 \\
Median Length & 9 \\
\bottomrule
\end{tabular}
\caption{Utterance length statistics, reflecting the conversational register of the source material.}
\label{tab:utterance_stats}
\end{table}

\paragraph{Subjectivity formulation.} We adopt binary subjectivity classification (\textsc{Subjective} vs.\ \textsc{Objective}) rather than fine-grained sentiment polarity or emotion, because subjectivity boundaries remain highly consistent across diverse dialects ($\kappa = 0.78$), whereas sentiment polarity is heavily influenced by dialect-specific emotional lexicalization.

\paragraph{Inter-annotator agreement.}
\label{subsec:iaa}
To validate dataset reliability, a stratified sample of $1{,}500$ entries ($300$ per dialect) was independently annotated by a second dialect expert, with disagreements resolved by a third senior linguist.
Categorical agreement yielded Cohen's $\kappa = 0.78$ for subjectivity, $\kappa = 0.74$ for Standard Bangla translation acceptance, and $\kappa = 0.71$ for English translation approval.
For free-form English translations, the two annotators' independent final edits converged at a mean ROUGE-L of $0.84$ ($0.79$--$0.91$ across dialects), demonstrating strong semantic consistency.
Full per-dialect agreement statistics and adjudication protocols are detailed in Appendix~\ref{app:iaa_details}.
\section{Experimental Setup}
\label{sec:method}

We benchmark large language models across three tasks defined over \name{}: \textbf{dialect-to-English translation}, \textbf{binary subjectivity classification}, and \textbf{dialect-to-Standard-Bangla normalization}.
Our experimental framework evaluates four orthogonal axes: prompting regime (zero-shot, few-shot, CoT), script representation (native Bangla vs.\ Romanized transliteration), model regime (closed-source vs.\ open-source), and adaptation regime (prompting vs.\ LoRA fine-tuning).

\paragraph{Task definitions.}
\begin{enumerate}
    \item \textbf{Dialect-to-English MT:} Given a dialectal utterance $x$, the model produces an English translation $\hat{y}$ scored against the human English reference $y$ in \name{}.
    \item \textbf{Subjectivity Classification:} The model assigns a binary label $\hat{s} \in \{\textsc{Subjective},\, \textsc{Objective}\}$ to the input utterance.
    \item \textbf{Dialect-to-Standard Normalization:} The model maps the dialectal utterance $x$ into its Standard (Cholito) Bangla equivalent $\hat{z}$, scored against the human-verified Standard Bangla reference $z$.
\end{enumerate}

\paragraph{Prompting strategies.} We evaluate three prompting regimes: (i)~\textbf{Zero-shot} prompting providing only task instructions and input; (ii)~\textbf{Few-shot} prompting supplying $36$ in-context demonstrations ($6$ per regional dialect plus Standard Bangla), resampled per query from the training partition; and (iii)~\textbf{Chain-of-Thought (CoT)} prompting~\cite{kojima2022large} eliciting an eight-step structured reasoning chain (dialect recognition, token parsing, transliteration, semantic interpretation, cultural analysis, English translation, subjectivity assessment, and consistency check) before generating the final answer. Full prompt templates are provided in Appendix~\ref{app:prompts}.

\paragraph{Script representation.} Every configuration is tested under two script representations: a \textit{native} condition where input is presented in original Bangla script, and a \textit{transliterated} condition where input is presented in Romanized transliteration. Target outputs and label spaces remain identical.

\paragraph{Parameter-efficient fine-tuning (LoRA).} For open-source models, we fine-tune separate LoRA adapters~\cite{hu2022lora} for each task-by-script combination using rank $r=16$, scaling $\alpha=32$, and dropout $0.05$ on attention projections. Closed-source models are excluded from this regime as weights are inaccessible. Full training hyperparameters, loss dynamics, and hardware specifications are detailed in Appendix~\ref{app:lora}.

\paragraph{Models and baselines.} We evaluate seven primary LLMs: closed-source models (Gemini~3~Flash~\cite{comanici2025gemini}, GPT-4o-mini~\cite{openai2024gpt4o}, Claude Haiku 4.5~\cite{anthropic2024claude}) and open-source models (Qwen-3-4B~\cite{qwen2024qwen25}, Gemma-4-4B~\cite{gemma2024gemma}, Llama-3.1-8B~\cite{meta2024llama3}, Mistral-7B~\cite{jiang2023mistral}). We additionally compare against dedicated Indic sequence-to-sequence baselines: IndicBART~\cite{dabre2022indicbart} and BanglaT5~\cite{bhattacharjee2023banglat5} fine-tuned on the identical training folds (Appendix~\ref{app:indic_baselines}).

\paragraph{Data splits.} All prompting experiments evaluate on a dialect-balanced subset of $\approx 100$ items per variety ($600$ total across the five dialects plus Standard Bangla). LoRA fine-tuning utilizes a disjoint dialect-balanced fold containing $160$ training instances and $40$ test instances per dialect ($1{,}000$ train / $200$ test total), with zero overlap across folds or prompting subsets.

\paragraph{Evaluation metrics.} To ensure robust and metric-independent assessment, translation is evaluated using surface $n$-gram metrics (BLEU~\cite{papineni2002bleu}, ROUGE-2, ROUGE-L~\cite{lin2004rouge}, METEOR~\cite{banerjee2005meteor}), character-level chrF++~\cite{popovic2015chrf}, and neural COMET~\cite{rei2020comet} (`wmt22-comet-da`). High absolute BLEU scores reflect the conversational register of short utterances (mean $11.2$ tokens) and formulaic phrases with unique English translations ($58.2\%$ exact string match under LoRA). Dialect-to-Standard normalization is evaluated via chrF++ and COMET against the human Standard Bangla reference. Subjectivity is evaluated using macro Precision, Recall, and F1.
\section{Results and Analysis}
\label{sec:results}

We benchmark seven LLMs across zero-shot and few-shot prompting (Table~\ref{tab:bench_zero_few}), chain-of-thought prompting (Table~\ref{tab:bench_cot}), and LoRA fine-tuning (Table~\ref{tab:bench_lora}) across translation, subjectivity, and dialect-to-standard normalization (Table~\ref{tab:bench_norm}).
All translation metrics are evaluated on English outputs against human references; normalization is scored via chrF++ and COMET against Standard Bangla; subjectivity is evaluated via macro Precision, Recall, and F1. All scores are macro-averaged across the five dialects plus Standard Bangla.

\begin{table*}[t]
\centering
\small
    \begin{tabular}{lccccccc}
        \toprule
        \multirow{2}{*}{Models}
        & \multicolumn{4}{c}{Translation}
        & \multicolumn{3}{c}{Subjectivity} \\
        \cmidrule(lr){2-5}\cmidrule(lr){6-8}
        & BLEU & R-2 & R-L & Meteor
        & P & R & F1 \\
        \midrule

        \multicolumn{8}{c}{\textit{Zero-Shot (Native Bangla)}} \\
        \midrule
        \multicolumn{8}{l}{\hspace{0.3em}\textit{Closed-Source LLMs}} \\
        \quad Gemini 3 Flash    & 46.4 & 54.0 & 71.7 & 71.5 & 76.5 & 70.1 & 69.6 \\
        \quad GPT-4o-mini       & 38.4 & 44.4 & 62.6 & 62.3 & 55.7 & 55.0 & 54.8 \\
        \quad Claude Haiku 4.5  & 38.0 & 43.6 & 61.2 & 61.4 & 73.8 & 67.1 & 66.2 \\
        \addlinespace
        \multicolumn{8}{l}{\hspace{0.3em}\textit{Open-Source LLMs}} \\
        \quad Qwen-3-4B         &  0.2 &  2.1 &  3.9 &  4.0 & 62.5 & 52.4 & 41.3 \\
        \quad Gemma-4-4B        & 31.9 & 37.6 & 56.6 & 55.7 & 72.3 & 65.8 & 65.1 \\
        \quad Llama-3.1-8B      & 42.5 & 48.9 & 65.6 & 65.7 & 71.5 & 71.6 & 70.6 \\
        \quad Mistral-7B        & 16.2 & 20.4 & 38.4 & 38.7 & 36.5 & 32.9 & 32.8 \\
        \midrule

        \multicolumn{8}{c}{\textit{Zero-Shot (Transliterated)}} \\
        \midrule
        \multicolumn{8}{l}{\hspace{0.3em}\textit{Closed-Source LLMs}} \\
        \quad Gemini 3 Flash    & 40.1 & 47.5 & 66.2 & 65.8 & 74.6 & 67.8 & 67.0 \\
        \quad GPT-4o-mini       & 26.4 & 32.2 & 51.0 & 50.8 & 54.6 & 54.5 & 54.3 \\
        \quad Claude Haiku 4.5  & 24.6 & 29.2 & 47.5 & 47.8 & 72.1 & 65.0 & 64.0 \\
        \addlinespace
        \multicolumn{8}{l}{\hspace{0.3em}\textit{Open-Source LLMs}} \\
        \quad Qwen-3-4B         &  0.0 &  1.6 &  2.7 &  2.8 & 54.2 & 52.8 & 42.4 \\
        \quad Gemma-4-4B        & 21.1 & 25.6 & 44.0 & 43.4 & 72.6 & 62.0 & 59.7 \\
        \quad Llama-3.1-8B      & 26.9 & 32.9 & 50.3 & 50.9 & 68.8 & 68.7 & 67.0 \\
        \quad Mistral-7B        &  7.8 &  9.7 & 24.0 & 24.9 & 37.3 & 33.6 & 34.8 \\
        \midrule

        \multicolumn{8}{c}{\textit{Few-Shot (Native Bangla)}} \\
        \midrule
        \multicolumn{8}{l}{\hspace{0.3em}\textit{Closed-Source LLMs}} \\
        \quad Gemini 3 Flash    & 44.9 & 53.7 & 71.0 & 72.4 & 75.9 & 66.1 & 63.9 \\
        \quad GPT-4o-mini       & 42.2 & 48.5 & 65.6 & 66.1 & 67.2 & 65.2 & 65.0 \\
        \quad Claude Haiku 4.5  & 41.0 & 48.2 & 65.2 & 65.5 & 74.9 & 62.9 & 59.0 \\
        \addlinespace
        \multicolumn{8}{l}{\hspace{0.3em}\textit{Open-Source LLMs}} \\
        \quad Qwen-3-4B         & 27.5 & 31.9 & 49.6 & 48.1 & 62.9 & 58.9 & 55.3 \\
        \quad Gemma-4-4B        & 30.7 & 39.2 & 57.4 & 57.9 & 71.5 & 67.1 & 66.3 \\
        \quad Llama-3.1-8B      & 33.2 & 40.5 & 57.3 & 58.2 & 62.2 & 61.8 & 61.5 \\
        \quad Mistral-7B        & 19.3 & 25.3 & 41.1 & 42.4 & 63.6 & 61.8 & 61.4 \\
        \midrule

        \multicolumn{8}{c}{\textit{Few-Shot (Transliterated)}} \\
        \midrule
        \multicolumn{8}{l}{\hspace{0.3em}\textit{Closed-Source LLMs}} \\
        \quad Gemini 3 Flash    & 36.5 & 39.0 & 61.0 & 60.0 & 73.9 & 64.1 & 61.9 \\
        \quad GPT-4o-mini       & 25.5 & 35.9 & 56.6 & 55.8 & 65.2 & 63.2 & 63.0 \\
        \quad Claude Haiku 4.5  & 29.9 & 32.6 & 55.4 & 52.5 & 72.9 & 60.9 & 57.0 \\
        \addlinespace
        \multicolumn{8}{l}{\hspace{0.3em}\textit{Open-Source LLMs}} \\
        \quad Qwen-3-4B         &  8.5 & 14.0 & 32.6 & 31.6 & 60.9 & 56.9 & 53.3 \\
        \quad Gemma-4-4B        & 28.1 & 32.1 & 54.1 & 54.1 & 69.5 & 65.1 & 64.3 \\
        \quad Llama-3.1-8B      & 13.8 & 16.2 & 37.1 & 38.3 & 60.2 & 59.8 & 59.5 \\
        \quad Mistral-7B        & 11.3 & 16.8 & 38.8 & 38.6 & 61.6 & 59.8 & 59.4 \\
        \bottomrule
    \end{tabular}
\caption{Zero-shot and few-shot benchmarking across translation and subjectivity classification under native and Romanized input. Scores are macro-averaged across the five dialects plus Standard Bangla.}
\label{tab:bench_zero_few}
\end{table*}

\begin{table*}[t]
\centering
\small
    \begin{tabular}{lccccccc}
        \toprule
        \multirow{2}{*}{Models}
        & \multicolumn{4}{c}{Translation}
        & \multicolumn{3}{c}{Subjectivity} \\
        \cmidrule(lr){2-5}\cmidrule(lr){6-8}
        & BLEU & R-2 & R-L & Meteor
        & P & R & F1 \\
        \midrule

        \multicolumn{8}{c}{\textit{Chain-of-Thought (Native Bangla)}} \\
        \midrule
        \multicolumn{8}{l}{\hspace{0.3em}\textit{Closed-Source LLMs}} \\
        \quad Gemini 3 Flash    & 38.2 & 47.4 & 66.4 & 67.7 & 75.4 & 68.9 & 68.4 \\
        \quad GPT-4o-mini       & 26.3 & 33.4 & 50.7 & 51.9 & 67.4 & 66.8 & 66.0 \\
        \quad Claude Haiku 4.5  & 25.4 & 34.6 & 51.9 & 54.3 & 72.9 & 68.0 & 68.0 \\
        \addlinespace
        \multicolumn{8}{l}{\hspace{0.3em}\textit{Open-Source LLMs}} \\
        \quad Qwen-3-4B         & 11.3 & 18.0 & 36.1 & 37.2 & 57.7 & 52.8 & 39.4 \\
        \quad Gemma-4-4B        & 24.0 & 32.0 & 51.1 & 51.4 & 72.7 & 71.3 & 71.5 \\
        \quad Llama-3.1-8B      & 33.6 & 40.6 & 58.0 & 58.5 & 71.2 & 70.4 & 69.9 \\
        \quad Mistral-7B        &  9.9 & 14.1 & 30.1 & 31.0 & 58.2 & 57.9 & 57.7 \\
        \midrule

        \multicolumn{8}{c}{\textit{Chain-of-Thought (Transliterated)}} \\
        \midrule
        \multicolumn{8}{l}{\hspace{0.3em}\textit{Closed-Source LLMs}} \\
        \quad Gemini 3 Flash    & 39.7 & 43.0 & 64.7 & 64.1 & 74.9 & 68.5 & 67.9 \\
        \quad GPT-4o-mini       & 26.8 & 36.0 & 53.6 & 53.0 & 64.5 & 63.3 & 64.0 \\
        \quad Claude Haiku 4.5  & 32.5 & 38.0 & 57.0 & 56.4 & 71.9 & 64.6 & 66.0 \\
        \addlinespace
        \multicolumn{8}{l}{\hspace{0.3em}\textit{Open-Source LLMs}} \\
        \quad Qwen-3-4B         &  6.4 & 11.4 & 25.8 & 27.1 & 56.4 & 52.2 & 41.5 \\
        \quad Gemma-4-4B        & 24.5 & 32.1 & 51.5 & 52.4 & 69.4 & 67.4 & 68.3 \\
        \quad Llama-3.1-8B      & 33.2 & 41.6 & 57.3 & 56.3 & 69.1 & 68.3 & 67.9 \\
        \quad Mistral-7B        & 11.6 & 17.6 & 34.6 & 35.0 & 58.6 & 57.4 & 58.1 \\
        \bottomrule
    \end{tabular}
\caption{Chain-of-thought benchmarking of LLMs across translation and subjectivity under native and Romanized input.}
\label{tab:bench_cot}
\end{table*}

\begin{table*}[t]
\centering
\small
    \begin{tabular}{lcccccccc}
        \toprule
        \multirow{2}{*}{Models}
        & \multicolumn{5}{c}{Translation}
        & \multicolumn{3}{c}{Subjectivity} \\
        \cmidrule(lr){2-6}\cmidrule(lr){7-9}
        & BLEU & R-L & chrF++ & COMET & Meteor
        & P & R & F1 \\
        \midrule

        \multicolumn{9}{c}{\textit{LoRA Fine-Tuning (Native Bangla)}} \\
        \midrule
        \multicolumn{9}{l}{\hspace{0.3em}\textit{Open-Source LLMs}} \\
        \quad Qwen-3-4B         & 65.9 & 78.6 & 74.67 & 87.72 & 79.1 & 77.2 & 77.7 & 77.3 \\
        \quad Gemma-4-4B        & 63.9 & 77.6 & 73.17 & 87.09 & 77.7 & 74.0 & 74.5 & 74.0 \\
        \quad Llama-3.1-8B      & 68.9 & 81.3 & 77.02 & 88.90 & 81.0 & 77.7 & 79.3 & 78.4 \\
        \quad Mistral-7B        & 73.6 & 84.7 & 80.43 & 89.97 & 84.2 & 77.8 & 79.3 & 78.3 \\
        \midrule

        \multicolumn{9}{c}{\textit{LoRA Fine-Tuning (Transliterated)}} \\
        \midrule
        \multicolumn{9}{l}{\hspace{0.3em}\textit{Open-Source LLMs}} \\
        \quad Qwen-3-4B         & 53.9 & 67.4 & 27.40 & 59.38 & 67.7 & 69.2 & 69.8 & 69.3 \\
        \quad Gemma-4-4B        & 51.9 & 66.3 & 30.45 & 62.15 & 66.3 & 66.1 & 67.3 & 66.0 \\
        \quad Llama-3.1-8B      & 56.9 & 69.3 & 30.63 & 62.09 & 69.0 & 69.9 & 71.5 & 70.2 \\
        \quad Mistral-7B        & 61.6 & 72.7 & 36.94 & 64.27 & 72.2 & 70.3 & 72.3 & 70.5 \\
        \bottomrule
    \end{tabular}
\caption{LoRA fine-tuning results on \name{} evaluated on the held-out test fold ($100$ per dialect; $600$ items including Standard Bangla). Closed-source models are omitted due to API weight restrictions. All four models outperform the strongest closed-source zero-shot baseline (Gemini 3 Flash: COMET $86.68$, chrF++ $63.67$).}
\label{tab:bench_lora}
\end{table*}

\begin{table}[!ht]
\centering
\resizebox{\columnwidth}{!}{%
    \begin{tabular}{lcccc}
        \toprule
        \multirow{2}{*}{Models} & \multicolumn{2}{c}{Zero-Shot (chrF++)} & \multicolumn{2}{c}{Few-Shot (chrF++)} \\
        \cmidrule(lr){2-3}\cmidrule(lr){4-5}
        & Native & Romanized & Native & Romanized \\
        \midrule
        \multicolumn{5}{l}{\textit{Closed-Source LLMs}} \\
        \quad Gemini 3 Flash    & \textbf{71.41} & \textbf{64.89} & \textbf{78.70} & \textbf{73.00} \\
        \quad GPT-4o-mini       & 58.10 & 47.36 & 63.58 & 54.77 \\
        \quad Claude Haiku 4.5  & 57.14 & 45.46 & 66.13 & 57.57 \\
        \addlinespace
        \multicolumn{5}{l}{\textit{Open-Source LLMs}} \\
        \quad Gemma-4-4B        & 53.05 & 44.75 & 56.71 & 48.17 \\
        \quad Mistral-7B        & 31.51 & 26.65 & 36.73 & 28.66 \\
        \quad Llama-3.1-8B      & 15.82$^*$ & 32.90 & 46.28 & 36.39 \\
        \quad Qwen-3-4B         & 38.96$^\dagger$ & 28.99 & 44.06 & 36.61 \\
        \midrule
        \quad \textit{Copy-input baseline} & 39.66 & --- & 39.66 & --- \\
        \bottomrule
    \end{tabular}%
}
\caption{Dialect-to-Standard-Bangla normalization macro chrF++ across zero-shot and few-shot prompting. $^*$Llama zero-shot native is depressed by $69.5\%$ output-script drift, repaired by few-shot. $^\dagger$Qwen-3-4B required constrained JSON decoding (\texttt{format="json"}): as a hybrid-reasoning model it emits reasoning text outside \texttt{<think>} delimiters and exhausts the generation budget before returning the required JSON object, yielding no parseable output under the unconstrained configuration used for all other models; constrained decoding governs output format only, not content quality. \emph{Copy-input baseline} is the score obtained by returning the dialectal input unchanged ($39.66$ chrF++, native script; near-zero for Romanized input against a Bangla-script reference, hence omitted), and marks the floor below which a system has not performed the task. Qwen-3-4B returns its input verbatim on $34.1\%$ of native items (all other models $1.8$--$11.1\%$), so its native-script scores are inflated by pass-through; on the items it does modify it reaches $46.25$ chrF++. Mistral-7B few-shot and Llama-3.1-8B zero-shot fall below the copy baseline in the native-script condition.}
\label{tab:bench_norm}
\end{table}

\subsection{Prompting Strategies vs.\ Model Competence}

A critical empirical insight is that \textbf{prompting-regime scores reflect the interaction between a model and a prompting strategy rather than its underlying Bangla competence}.
While closed-source frontier models maintain steady performance under zero-shot prompting (Gemini leading at $46.4$ BLEU / $71.41$ normalization chrF++), open-source models suffer severe strategy-specific pathology:
\begin{itemize}
    \item \textbf{Llama-3.1-8B Demonstration Interference:} Llama degrades sharply under few-shot translation ($42.5 \to 33.2$ BLEU). Item-level error auditing reveals \textit{source neglect}: correct zero-shot outputs are replaced under few-shot by fluent demonstration-register sentences unrelated to the source utterance. Llama's correct-to-unrelated flip rate is $5.9\%$ (a $15\times$--$30\times$ outlier compared to $0.2$--$0.4\%$ for all other models; Table~\ref{tab:app_llama_flips}). A prompt-count sweep (Appendix~\ref{app:prompt_sweep}) confirms this interference is regime-triggered at even $1$ demo/dialect ($38.6$ chrF++ vs.\ $54.6$ zero-shot). In normalization, Llama zero-shot exhibits $69.5\%$ output-script drift (emitting Romanized text with diacritics instead of Bangla script), which few-shot prompting repairs to $0\%$.
    \item \textbf{Qwen-3-4B Format Collapse:} Qwen's near-zero zero-shot scores ($0.2$ BLEU) are caused by a $91.4\%$ unparseable generation/format collapse rather than poor translation capability; few-shot and CoT prompting repair formatting, restoring scores to $27.5$ and $11.3$ BLEU.
    \item \textbf{CoT Task Asymmetry:} CoT consistently improves categorical subjectivity classification (e.g., GPT-4o-mini $+11.2$ F1, Gemma $+6.4$ F1) by encouraging structured pragmatic analysis. However, CoT degrades translation surface fidelity across all closed models (Gemini $46.4 \to 38.2$ BLEU). Error decomposition (Appendix~\ref{app:cot_decomp}) reveals this drop is driven by \textit{paraphrase drift} and verbosity (producing $45.2\%$ CompleteMistranslation under surface metrics), while neural COMET drops much less ($\Delta\text{COMET} = 3.4$ vs.\ $\Delta\text{chrF++} = 5.8$).
\end{itemize}
Because prompting strategies introduce confounding interactions, we treat \textbf{LoRA fine-tuning as the primary measure of model capacity}.

\subsection{Causal Decomposition of the Transliteration Penalty}
\label{subsec:script_effects}

Across all seven LLMs, three tasks, and four regimes (Tables~\ref{tab:bench_zero_few}--\ref{tab:bench_norm}), \textbf{Romanized input consistently underperforms native Bangla script}.
Every finding is metric-robust: re-scoring all $35{,}460$ outputs with neural COMET (`wmt22-comet-da`) and chrF++ confirms that native script outperforms Romanized input in all $19$ model-regime pairs.
We isolate the mechanism behind this penalty into three causal components:

\paragraph{1. Many-to-One Information Loss.} Standard Romanization (Avro) collapses distinct phonemic features (vowel length, aspiration, and nasal distinctions; Figure~\ref{fig:app_romanization}).
To quantify this, we executed a \textit{round-trip inverse transliteration control} (Appendix~\ref{app:roundtrip}) by mapping Romanized text back to native script via deterministic inverse mapping.
Character recovery is poor (mean CER $0.14$--$0.21$), and re-evaluating LLMs on back-transliterated text recovers essentially \textbf{none} of the performance gap (Gemini zero-shot chrF++: native $62.13$, round-trip $55.57$, Romanized $57.81$; Mistral: native $29.35$, round-trip $22.52$, Romanized $23.17$).
The penalty is dominated by irreversible information destroyed in the mapping channel.

\paragraph{2. Subword Token Fragmentation.} In aggregate, Romanized text is more token-compact than native Bangla ($\approx 0.4$ vs.\ $0.4$--$1.25$ tokens/char), ruling out a simple context-budget inefficiency.
However, an item-level regression of the native-minus-Romanized COMET gap on token inflation (Appendix~\ref{app:token_regression}) is positive and statistically significant for all four open models under LoRA ($\beta = +7.86$ to $+60.92, p \le 0.035$).
Atypical subword fragmentation serves as the per-item signature of damaged, ambiguous mappings.

\paragraph{3. Distributional Sparsity and Scheme Robustness.} Unstandardized Romanization scatters probability mass across spelling variants.
To verify that this is not an artifact of the Avro convention, we benchmarked models under two alternative Romanization schemes: ITRANS and ISO-15919 (Appendix~\ref{app:scheme_robustness}).
The penalty holds across all schemes, and Avro is the \textit{mildest} penalty (Gemini penalty: Avro $4.63$ vs.\ ITRANS $8.92$ vs.\ ISO $4.24$ chrF++), proving that our reported gaps are conservative lower bounds.

\begin{figure*}[!t]
\centering
\includegraphics[width=0.95\textwidth]{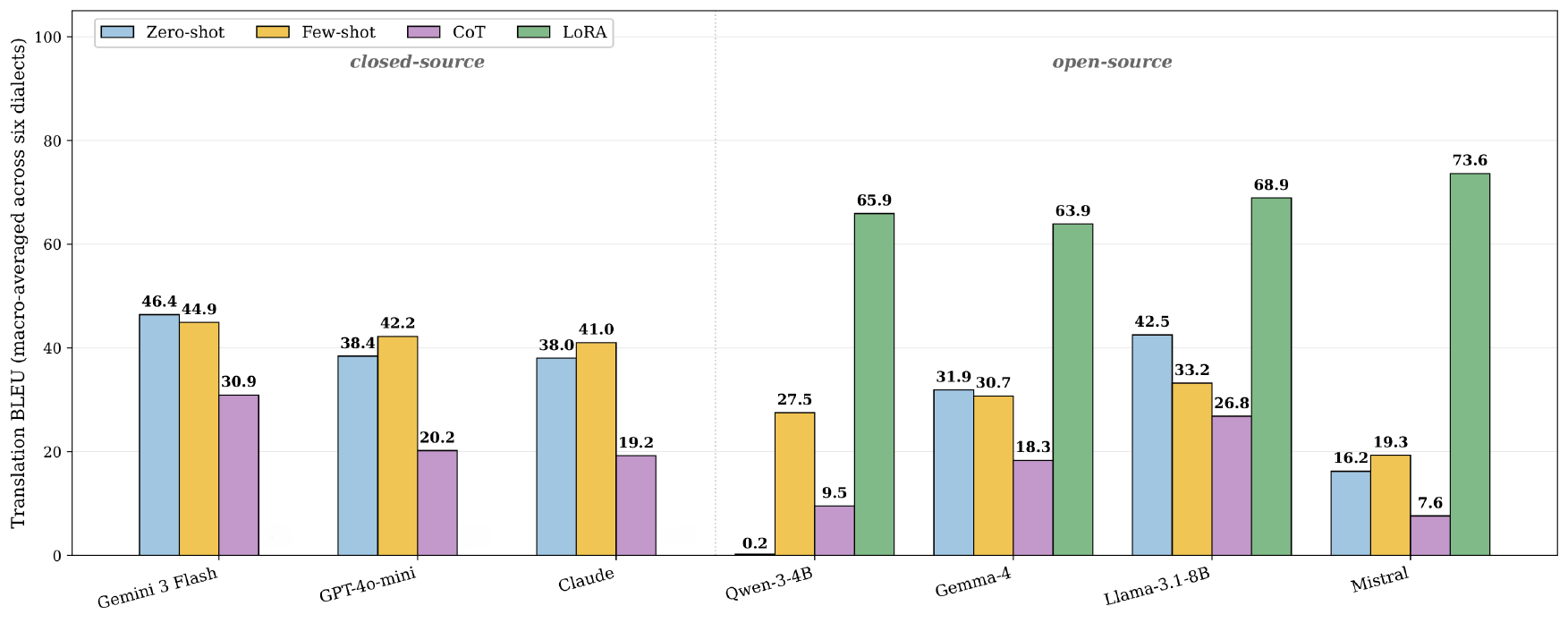}
\caption{Translation BLEU across four regimes (zero-shot, few-shot, chain-of-thought, LoRA fine-tuning) for all seven models, native Bangla input, macro-averaged across dialects. LoRA elevates all four open-source models into a $64$--$74$ BLEU band, with Mistral-7B gaining $+57.4$ BLEU over its zero-shot baseline and surpassing all closed-source models.}
\label{fig:bleu_regimes}
\end{figure*}

\subsection{Supervision and Adaptation Dynamics}

\begin{figure*}[!t]
\centering
\includegraphics[width=0.95\textwidth]{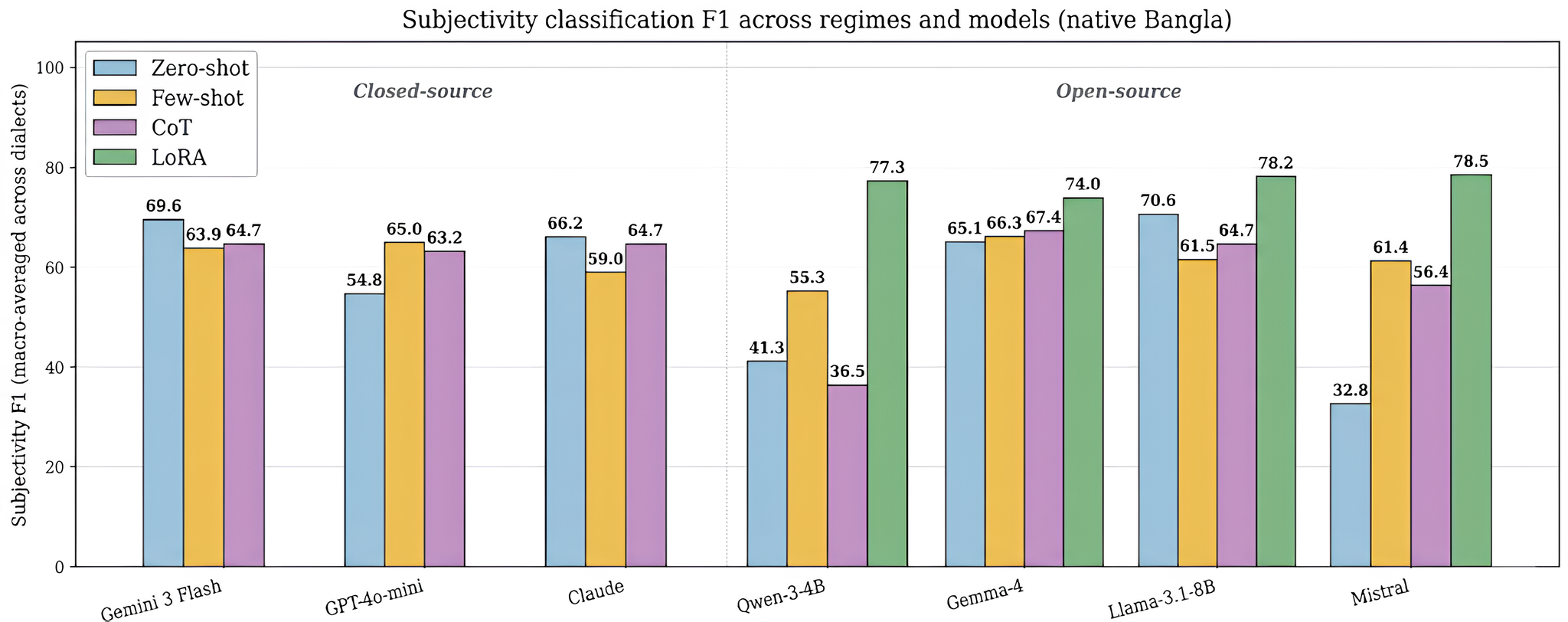}
\caption{Subjectivity classification F1 across four regimes for all seven models (native script). LoRA lifts every open-source model into the $74$--$79$ F1 band, closing the open-vs-closed performance gap.}
\label{fig:app_subj_f1_grouped}
\end{figure*}

Fine-tuning open-source models via LoRA on only $160$ examples per dialect yields large gains on both benchmarked generation tasks---translation and subjectivity classification (Table~\ref{tab:bench_lora}, Figures~\ref{fig:bleu_regimes}--\ref{fig:app_subj_f1_grouped}). We did not fine-tune adapters for normalization; the normalization results in Table~\ref{tab:bench_norm} are prompting-only, and the supervised comparison available for that task is the Indic seq2seq fine-tune in Appendix~\ref{app:indic_baselines}:
\begin{itemize}
    \item \textbf{Translation:} Mistral-7B jumps from $16.2$ to $73.6$ BLEU (COMET $89.97$), outperforming Gemini~3~Flash zero-shot ($46.4$ BLEU / COMET $86.68$). Gemma-4-4B, Qwen-3-4B, and Llama-3.1-8B reach $63.9$--$68.9$ BLEU.
    \item \textbf{Subjectivity:} All four open models land in the $74.0$--$78.4$ F1 range, exceeding all closed-source zero-shot and few-shot baselines.
    \item \textbf{Comparison to Indic Baselines:} Dedicated seq2seq models fine-tuned on the identical $160$/dialect fold trail LoRA-adapted LLMs substantially: IndicBART achieves $11.19$ chrF++ ($7.39$ BLEU) on translation---a figure dominated by its $50.6\%$ empty-output rate, rising to $22.66$ chrF++ on the items it does generate---and $40.82$ chrF++ on normalization; BanglaT5 reaches $42.06$ chrF++ ($27.74$ BLEU) on translation and $44.47$ on normalization (Appendix~\ref{app:indic_baselines}).
    \item \textbf{Persistent Script Gap:} Even under LoRA, the native-over-transliterated gap remains substantial ($\Delta = 12.0$--$17.0$ BLEU; $\beta = 38$--$61, p < 0.001$), confirming that $160$ examples cannot reconstruct distinctions destroyed in the Romanization channel.
\end{itemize}

\subsection{Linguistic Divergence and Error Analysis}
\label{subsec:error_analysis}

A feature-conditioned error analysis (Appendix~\ref{app:error_taxonomy}) reveals that error rates mirror linguistic distance from Standard Bangla:
\begin{itemize}
    \item \textbf{Lexical vs.\ Phonological Divergence:} High-lexical-divergence items exhibit $\approx 2\times$ the \textsc{CompleteMistranslation} rate of low-divergence items zero-shot ($27.4\%$ vs.\ $13.4\%$) and remain resistant even after LoRA ($9.8\%$ vs.\ $5.9\%$). Specifically, Chittagong non-cognate \emph{goijja}-class verbs remain hardest after adaptation ($16.2\%$ mistranslation, chrF++ $54.4$), whereas Barisal phonological $h^C$-cluster shifts are fully mastered ($0\%$ error, chrF++ $92.3$).
    \item \textbf{Error Taxonomy Progression:} LoRA shifts the prediction distribution toward \textsc{Success} ($17.2\% \to 58.2\%$) and away from \textsc{CompleteMistranslation} ($29.1\% \to 9.5\%$), while CoT produces the highest mistranslation rate ($45.2\%$) due to unconstrained paraphrase generation.
\end{itemize}
Detailed per-dialect tables, confusion matrices, and qualitative failure case studies are provided in Appendices~\ref{app:full_results}--\ref{app:case_studies}.

\section{Conclusion}
\label{sec:conclusion}
We introduce \name, the first Bangla dialect benchmark to align Romanized transliteration as a factorial variable with Standard Bangla, English, and subjectivity labels across five regional dialects ($6{,}000$ entries). Experiments on seven LLMs show that LoRA fine-tuning with only $160$ examples per dialect surpasses closed-source models, Romanized transliteration consistently degrades current LLMs due to many-to-one information loss and token fragmentation, and dialect-aligned supervision is the critical bottleneck for Bangla dialect understanding.

\section*{Limitations}

\name{} has two primary limitations. First, the dataset is naturally imbalanced across dialects ($1{,}900$ Chittagong entries vs.\ $700$ each for Barisal and Rangpur), reflecting the differential online availability of dialect-tagged content; while our evaluation protocols sample dialect-balanced subsets to prevent the imbalance from confounding macro-averaged scores, the imbalance may still bias future cross-dialect transfer studies. Second, Bangla dialects lack fixed transliteration rules or a standardized Romanization grammar: while our multi-scheme evaluations confirm that the transliteration penalty is robust across Avro, ITRANS, and ISO-15919 conventions (Appendix~\ref{app:tokenizer_decoding}), downstream users should account for orthographic variance when comparing across transliterated resources. Beyond these, our LoRA experiments evaluate a $160$-instance-per-dialect configuration ($1{,}000$ train total), representing a practical lower bound on what adaptation can achieve; while we benchmark fine-tuned Indic seq2seq models (IndicBART and BanglaT5; Appendix~\ref{app:indic_baselines}), larger Indic foundation models remain future work; and our dataset captures written online dialectal Bangla, which may differ from spoken dialectal varieties.

\paragraph{Scope of the reported experiments.} Several analyses that would further constrain our claims are outside the scope of this version. Our parameter-efficient fine-tuning results rest on a single training budget ($160$ examples per dialect) and one hyperparameter configuration; we do not report a data-scaling curve, so the claim that dialect-aligned supervision is the binding constraint is supported by a single-budget comparison against prompting rather than by a trend. Our adaptation experiments cover translation and subjectivity classification; dialect-to-standard normalization is benchmarked under prompting only, and the sole supervised normalization comparison we report is the Indic seq2seq fine-tune in Appendix~\ref{app:indic_baselines}, which suggests that $160$ dialect-aligned examples are not sufficient for a small Indic seq2seq model to match strong prompting on this task; the extent to which LoRA transfers the translation-side gains to normalization remains open. The demonstration-count sweep (Appendix~\ref{app:prompt_sweep}) covers Llama-3.1-8B, Mistral-7B and Gemma-4-4B; Qwen-3-4B is excluded because it requires constrained JSON decoding (Table~\ref{tab:bench_norm}), which would confound a cross-model comparison, and the closed-source models are excluded on API cost. Automatic evaluation is metric-diverse---BLEU, chrF++, and COMET across all $35{,}460$ outputs---but we do not report human adequacy or fluency judgements, nor a rubric-based audit of chain-of-thought intermediate steps; both remain future work. Finally, we report per-dialect inter-annotator agreement (Table~\ref{tab:app_iaa_per_dialect}) and the overall adjudication rejection rate, but not per-dialect rejection rates, which were not logged separately during annotation.

\section*{Ethical Considerations}

All data was collected from publicly accessible online sources; no private, personally identifying, or otherwise sensitive information was deliberately collected. The dataset contains only ordinary user-generated sentences (everyday conversations, opinions, and commentary) and does not include private communication, personal correspondence, financial or medical information, government identifiers, or other sensitive content. Where user identifiers or contact information incidentally appeared in source text, they were removed during the verification stage. The released dataset poses no privacy risk to the original posters of the source content. All annotation was performed in-house by the research team and contracted student annotators, who were briefed on data handling and confidentiality and paid above the local research-assistantship rate. \name{} is intended for research on dialect-aware NLP, low-resource language technology, dialect identification, dialect-to-standard normalization, and cross-dialectal transfer; it is \emph{not} intended for inferring demographic or geographic attributes of individual users, and should not be used as a basis for any such inference.

\subsection*{Usage of AI}
The authors employed AI tools solely for limited language polishing and grammatical improvements in selected sections of this manuscript. All scientific aspects of the work, including study design, data acquisition, annotation, analysis, interpretation, and conclusions, were independently carried out by the authors. No AI system was used to generate findings, create figures or tables, review or synthesize literature, or formulate scientific claims. The authors assume full responsibility for the accuracy and integrity of the content presented in this paper.


\bibliography{anthology}

\appendix

\setcounter{topnumber}{4}
\setcounter{bottomnumber}{3}
\setcounter{totalnumber}{6}
\setcounter{dbltopnumber}{4}
\renewcommand{\topfraction}{0.92}
\renewcommand{\bottomfraction}{0.9}
\renewcommand{\textfraction}{0.07}
\renewcommand{\floatpagefraction}{0.75}
\renewcommand{\dbltopfraction}{0.92}
\renewcommand{\dblfloatpagefraction}{0.75}

\newcommand{\promptbox}[2][\columnwidth]{%
  \fbox{\parbox{\dimexpr#1-2\fboxsep-2\fboxrule\relax}{%
    \small\ttfamily\raggedright\setlength{\parindent}{0pt}#2}}%
}

\section{Dialect Linguistic Background}
\label{app:dialect_background}

Bangla regional dialects differ from Standard Bangla along phonological, lexical, and morphological axes that are systematic enough to be diagnostic on a per-sentence basis but variable enough to resist simple rule-based normalization. The five dialects in \name{} span the major regional varieties of Bangladesh and exhibit the following signatures:

\paragraph{Chittagong (Chattogram).} Spoken across the southeastern divisions of Bangladesh, Chittagong shows the strongest divergence from Standard Bangla of the five dialects in our benchmark, to the point that mutual intelligibility with the standard is partial rather than full. Diagnostic features include the dialect-final particle \emph{dey} as an interrogative or assertive marker (e.g., \emph{kemne goijji dey}, ``how have you done this?''), retroflex consonant shifts (\emph{r} $\to$ \emph{lr}, \emph{l} $\to$ \emph{r} in onset positions), the locative postposition \emph{tu}/\emph{at} in place of standard \emph{ke}/\emph{te}, and extensive lexical replacement of standard verbs (e.g., \emph{goijja} for \emph{kora}, ``to do''). Chittagong has the lowest per-dialect BLEU under every regime in our experiments and the largest residual error rate even after LoRA adaptation; we attribute this to the combined effect of high lexical divergence and lower pretraining coverage compared to standard-adjacent varieties.

\paragraph{Barisal (Barishal).} A coastal southern dialect with comparatively mild divergence from Standard Bangla. Diagnostic features include the consonant cluster simplification \emph{r}\textsuperscript{C} $\to$ \emph{h}\textsuperscript{C} (\emph{karchen} $\to$ \emph{horlen}, ``did''), vowel raising in word-medial positions, and adoption of the present-perfect auxiliary \emph{ase} where Standard Bangla would prefer the simple past. Barisal sentences are usually intelligible to a Standard Bangla speaker after a few exposures and our LoRA-tuned models reach near-saturation ($85$+ BLEU) on this dialect.

\paragraph{Noakhali.} Geographically adjacent to Chittagong but linguistically a distinct continuum, Noakhali shows interior shifts in vowel quality (notably the front-vowel reduction \emph{e} $\to$ \emph{ya} word-initially, \emph{aitchi} from \emph{esechi}, ``I have come''), the use of dialect-specific possessive pronouns (\emph{annei}/\emph{anne}, ``yours''), and frequent consonant elision in postpositions. Noakhali sits between Chittagong and Barisal in benchmark difficulty.

\paragraph{Sylhet.} A northeastern dialect with extensive contact-influence from Assamese and Sylheti Nagri orthographic traditions. Diagnostic features include retroflex-flap substitutions (\emph{kh}/\emph{r} alternations), distinctive question particles (\emph{kita}, ``what''), and a verbal aspect system that distinguishes habitual and progressive in surface forms that Standard Bangla collapses. Sylhet's lexicon contains a substantial number of items with no transparent Standard Bangla cognate, which manifests in our error analysis as a higher \textsc{CompleteMistranslation} rate even after fine-tuning.

\paragraph{Rangpur.} A northwestern dialect spoken across the Rangpur and Rajshahi divisions. Diagnostic features include rising-tone lengthening on stressed vowels, the second-person plural \emph{tomara} $\to$ \emph{tora}/\emph{tomra} alternation, dialect-specific quantifiers (\emph{kuna} for \emph{kichu}, ``some''), and a future-tense suffix that surfaces as \emph{-im}/\emph{-iya} rather than standard \emph{-bo}. Rangpur is the closest of the five dialects to Standard Bangla in lexical overlap and produces the strongest zero-shot baseline.

\paragraph{Comparison with DIALTSA-BN.} Table~\ref{tab:app_dialtsa_comp} provides an explicit side-by-side comparison between \name{} and the prior benchmark DIALTSA-BN~\cite{jawad2025dialtsabn}. \name{} expands the dataset scale ten-fold, covers five dialects (adding Rangpur), provides five fully aligned fields, factorially evaluates script effects across four regimes, and incorporates causal tokenizer and failure analyses.

\begin{table*}[!t]
\centering
\small
\begin{tabular}{>{\raggedright\arraybackslash}p{3.1cm}>{\raggedright\arraybackslash}p{5.0cm}>{\raggedright\arraybackslash}p{6.0cm}}
\toprule
\textbf{Dimension} & \textbf{DIALTSA-BN}~\cite{jawad2025dialtsabn} & \textbf{\name{} (Ours)} \\
\midrule
Dataset Scale & $600$ utterances & $6{,}000$ utterances ($10\times$ expansion) \\
Dialect Coverage & $4$ regional dialects & $5$ regional dialects (adding Rangpur) \\
Aligned Fields per Entry & $3$ fields (dialect, Standard Bangla, sentiment) & $5$ fields (dialect, Romanization, Standard Bangla, English, subjectivity) \\
Script Evaluation & Uncontrolled / ad-hoc & Factorial native vs.\ Romanized across all models and regimes \\
Adaptation Regimes & Prompting-only (zero-shot, few-shot) & Zero-shot, few-shot, CoT, and LoRA parameter-efficient fine-tuning \\
Mechanistic \& Causal Analysis & None (descriptive reporting) & Tokenizer fertility regression, round-trip control, multi-scheme robustness, failure diagnostics \\
\bottomrule
\end{tabular}
\caption{Systematic comparison between DIALTSA-BN and \name.}
\label{tab:app_dialtsa_comp}
\end{table*}

\section{Data Collection and Source Filtering}
\label{app:data_collection}

\paragraph{Source platforms.} Candidate utterances were drawn from publicly accessible content on YouTube (regional-comedy and regional-news comment threads, including channels associated with Bangladesh Television's regional service BTV Chittagong, language-documentation channels such as \emph{Indo-Aryan: Chittagonian} and \emph{I Love Languages}, and dialect-comedy creators on short-form video platforms), Facebook (open public groups dedicated to regional Bangladeshi communities and dialect-themed pages associated with Sylheti-revival movements such as the Greater Sylot Society and the Sylheti Project), Reddit (subreddits such as \emph{r/bangladesh}, \emph{r/dhaka}, and regional discussion threads), Bangla regional blogs and online opinion-and-feature articles (The Daily Star's ``Chittagonian humour'' column, The Business Standard's Sylheti-preservation features, and divisional desks at Bangla Tribune, Dhaka Post, and Ajker Patrika that publish in regional-flavored Bangla), and dialect-tagged open-source corpora released on Kaggle and Mendeley Data (Vashantor~\cite{faria2023vashantor}, ONUBAD~\cite{sultana2025onubad}, ChatgaiyyaAlap~\cite{chowdhury2025chatgaiyyaalap}, ANCHOLIK-NER~\cite{paul2025ancholik}, and BanglaDial). A representative URL inventory of the source platforms is reproduced in Table~\ref{tab:app_sources}. We deliberately excluded private communications, paywalled content, and any platform requiring user authentication beyond a public read scope.

\begin{table*}[!t]
\centering
\footnotesize
\begin{tabular}{>{\raggedright\arraybackslash}p{2.2cm}>{\raggedright\arraybackslash}p{3.7cm}>{\raggedright\arraybackslash}p{7.2cm}}
\toprule
Category & Platform / Source & Reference URL or DOI \\
\midrule
\multirow{3}{*}{\makecell[l]{Reference \\corpora}}
& Vashantor & cited as \citet{faria2023vashantor} \\
& ONUBAD & DOI \texttt{10.17632/k769s4vk5z.2} \citep{sultana2025onubad} \\
& BanglaDial & PubMed Central PMC12597015 \\
\midrule
\multirow{3}{*}{\makecell[l]{Wikipedia / \\encyclopedic}}
& Chittagonian language & \href{https://en.wikipedia.org/wiki/Chittagonian_language}{\texttt{en.wikipedia.org/wiki/Chittagonian\_language}} \\
& Sylheti language & \href{https://en.wikipedia.org/wiki/Sylheti_language}{\texttt{en.wikipedia.org/wiki/Sylheti\_language}} \\
& BTV Chittagong & \href{https://en.wikipedia.org/wiki/BTV_Chittagong}{\texttt{en.wikipedia.org/wiki/BTV\_Chittagong}} \\
\midrule
\multirow{3}{*}{\makecell[l]{News and \\feature articles}}
& The Daily Star (Chittagonian humour) & \url{thedailystar.net/news/chittagonian-humour} \\
& The Business Standard (Sylheti preservation) & \href{https://tbsnews.net/features/panorama/effort-preserve-heritage-sylheti-language-1098446}{\texttt{tbsnews.net (article 1098446)}} \\
& Bangla Tribune / Dhaka Post (Rangpur) & \url{banglatribune.com/rangpur-news}, \url{dhakapost.com/country/rangpur-news} \\
\midrule
\multirow{2}{*}{\makecell[l]{Curated \\dataset hubs}}
& Bangla NLP datasets (Foysal87) & \url{github.com/Foysal87/Bangla-NLP-Dataset} \\
& ANCHOLIK-NER & arXiv \texttt{2502.11198} \citep{paul2025ancholik} \\
\bottomrule
\end{tabular}
\caption{Representative source platforms and reference URLs used during the harvesting and authenticity-verification stages of \name. We list publicly accessible sources only; private group content, paywalled articles, and authenticated platforms were excluded. The full per-platform URL inventory for each dialect is released alongside the dataset.}
\label{tab:app_sources}
\end{table*}

\paragraph{Initial harvest and filtering.} The raw harvest yielded approximately $24{,}000$ candidate utterances across the five target dialects. We applied a four-stage filtering pipeline:
\begin{enumerate}
\item \textbf{Language identification.} Candidate utterances were passed through a Bangla language identifier with a minimum confidence threshold of $0.85$. Non-Bangla candidates (English code-mixed segments dominated by English, Hindi loan-paragraphs) were discarded.
\item \textbf{Length filter.} Utterances shorter than three tokens or longer than $40$ tokens were dropped. Very short utterances tend to be acknowledgement tokens (e.g., ``ok'', ``hmm'') that carry no dialectal signal; very long utterances tend to be code-mixed multi-sentence comments that complicate per-sentence dialect attribution.
\item \textbf{Native-speaker verification.} A native speaker of each dialect reviewed all candidates pre-tagged with that dialect's region under Guideline G1 (Appendix~\ref{app:guidelines}). The reviewer confirmed that the utterance was a legitimate instance of the target dialect (not standard Bangla in regional spelling), that it contained at least one dialect-diagnostic feature, and that it was not heavily code-mixed with English. Rejection rates at this stage averaged $34\%$ across dialects ($46\%$ for Barisal, $41\%$ for Rangpur, $35\%$ for Noakhali, $28\%$ for Sylhet, and $20\%$ for Chittagong).
\item \textbf{Deduplication.} Exact-match and near-match (Levenshtein-ratio $> 0.9$) deduplication was applied within each dialect.
\end{enumerate}

After filtering, $6{,}000$ utterances remained for annotation, distributed as shown in Table~\ref{tab:app_dialect_dist}. The distribution reflects natural online availability across platforms.

\begin{table*}[!t]
\centering
\small
\begin{tabular}{lcccccc}
\toprule
\textbf{Dialect} & \textbf{YouTube} & \textbf{Facebook} & \textbf{Reddit} & \textbf{News/Blogs} & \textbf{Ref.\ Corpora} & \textbf{Total Entries} \\
\midrule
Chittagong & 850 & 420 & 180 & 210 & 240 & 1{,}900 \\
Noakhali   & 680 & 360 & 140 & 160 & 160 & 1{,}500 \\
Sylhet     & 510 & 280 & 120 & 140 & 150 & 1{,}200 \\
Barisal    & 310 & 160 & 60 & 80 & 90 & 700 \\
Rangpur    & 290 & 180 & 70 & 80 & 80 & 700 \\
\midrule
\textbf{Total} & \textbf{2{,}640} & \textbf{1{,}400} & \textbf{570} & \textbf{670} & \textbf{720} & \textbf{6{,}000} \\
\bottomrule
\end{tabular}
\caption{Detailed dialect and source platform distribution of the $6{,}000$ curated utterances in \name, broken down by newly harvested online sources and seed reference corpora.}
\label{tab:app_dialect_dist}
\end{table*}

\section{Annotation Pipeline}
\label{app:annotation_pipeline}

\begin{figure*}[!t]
    \centering
    \includegraphics[width=\textwidth]{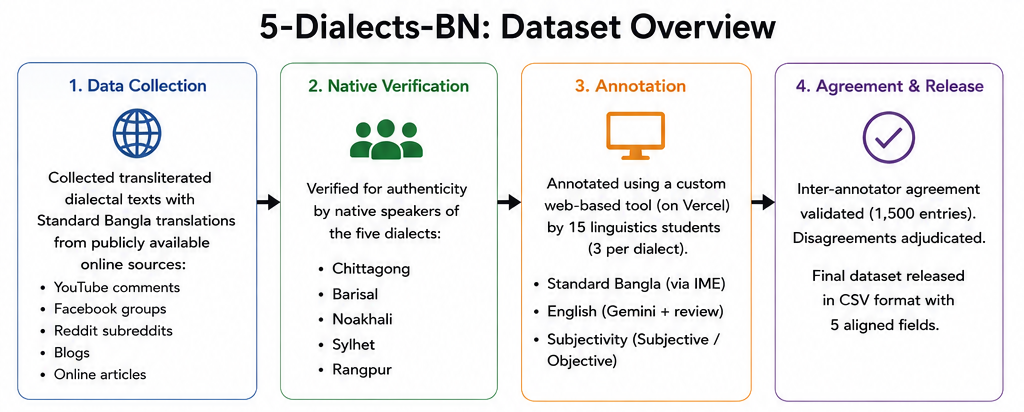}
    \caption{Construction workflow of \name. Transliterated dialectal texts with their Standard Bangla translations are collected from publicly available online sources, verified for authenticity by native speakers of each of the five target dialects (Chittagong, Barisal, Noakhali, Sylhet, Rangpur), and then enriched through a custom web-based annotation tool by fifteen undergraduate annotators from a Bangla Linguistics department. The final dataset is released in CSV format with five aligned fields per entry.}
    \label{fig:dataset_pipeline}
\end{figure*}

Figure~\ref{fig:dataset_pipeline} summarizes the four-stage construction workflow of \name{}: harvesting, native-speaker verification, tool-assisted annotation, and agreement-checked release.

\begin{figure*}[!t]
\centering
\includegraphics[width=0.95\textwidth]{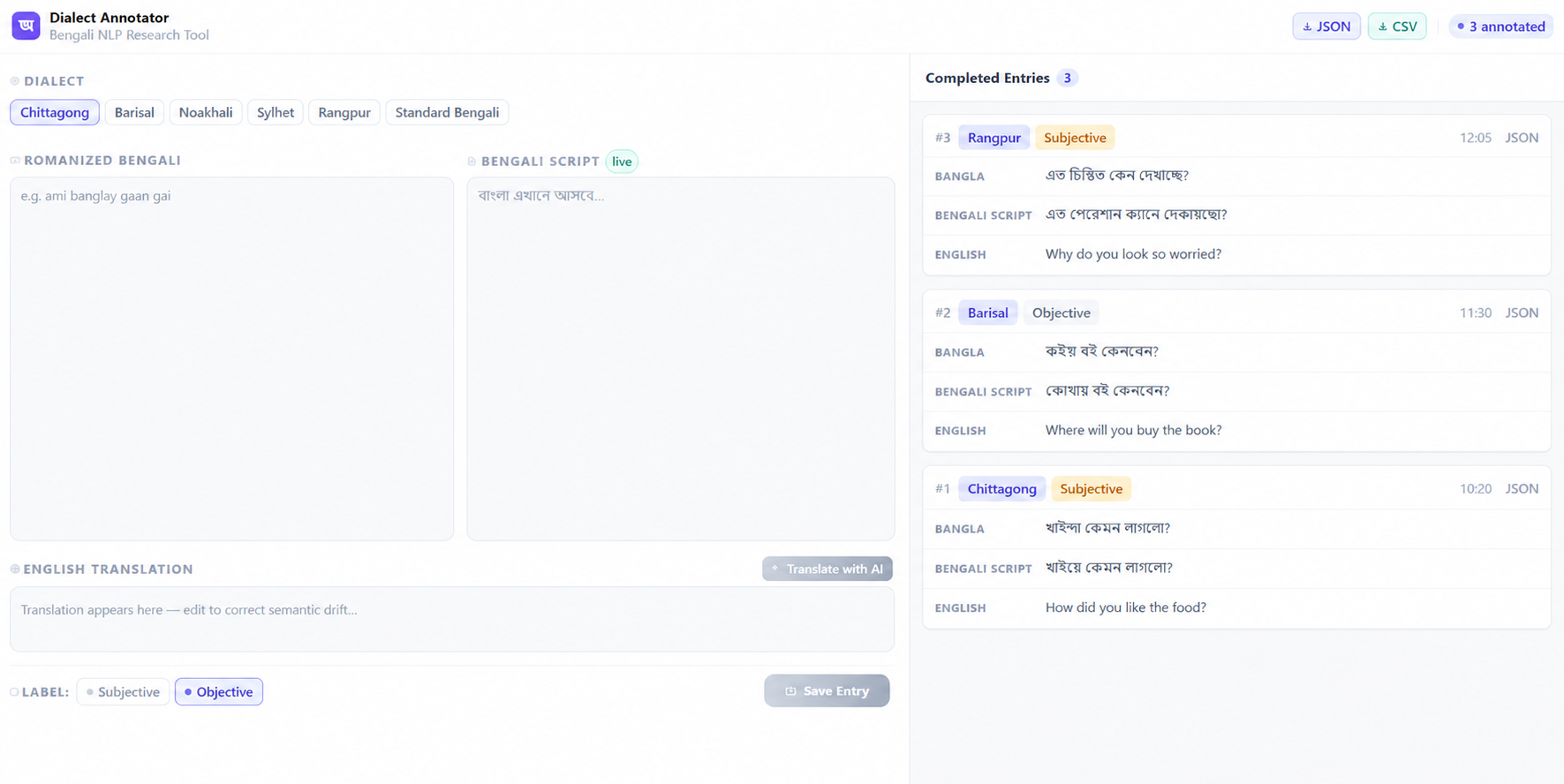}
\caption{Production interface of the \name{} annotation tool. Live tool hosted at \url{https://bangla-dialect-annotator.vercel.app/}.}
\label{fig:annotation_ui}
\end{figure*}

\paragraph{Tool architecture.} We developed a custom web-based annotation tool deployed on Vercel as a React/Next.js single-page application backed by a serverless storage layer (Figure~\ref{fig:annotation_ui}; live at \url{https://bangla-dialect-annotator.vercel.app/}). The interface presents a single annotation row at a time with the following workflow stages:
\begin{enumerate}
\item \textbf{Dialect selection.} The annotator selects their assigned dialect from a dropdown.
\item \textbf{Side-by-side preview.} The tool displays the Romanized dialectal text and the candidate Standard Bangla translation as parallel reading panels.
\item \textbf{Avro input method integration.} The annotator types or pastes the Romanized form into an integrated Avro IM editor, which converts Romanized Bangla to native Bangla script in real time. The annotator edits the native rendering inline to correct conversion errors.
\item \textbf{Standard Bangla quality check.} The annotator confirms that the Standard Bangla version preserves the meaning of the dialectal utterance, editing if necessary.
\item \textbf{Candidate English generation.} The tool invokes Gemini, conditioned on the verified Standard Bangla text (not the noisier transliterated form), to produce a candidate English translation.
\item \textbf{English quality check.} The annotator reviews the candidate English translation, edits it to correct semantic drift, and accepts the final version ($41\%$ edited overall; $52\%$ Sylhet, $48\%$ Chittagong, $35\%$ Barisal, $31\%$ Rangpur).
\item \textbf{Subjectivity labeling.} A binary checkbox interface offers \textsc{Subjective} and \textsc{Objective} as mutually exclusive choices.
\item \textbf{Auto-advance.} The next row loads automatically upon commit.
\end{enumerate}

\paragraph{Annotator pool.} Fifteen annotators were recruited from a Bangla Linguistics department, three per dialect, all native or proficient speakers of one or more target dialects. Annotators received a two-hour structured training on the tool and guidelines (Appendix~\ref{app:guidelines}) and were paid strictly above the local research-assistantship rate.

\paragraph{Adjudication protocol.} Disagreement on the $1{,}500$-entry double-annotated subset was resolved by a third annotator from the same dialect pool, blind to the original two annotations.

\section{Annotation Guidelines}
\label{app:guidelines}

\subsection{Guideline G1: Dialectal Authenticity}
Accept the utterance as a valid instance of the target dialect only if it satisfies at least one of: (i)~contains a diagnostic dialect-specific lexical item; (ii)~shows a diagnostic morphological feature of the target dialect (verbal suffix, postposition, pronoun); (iii)~shows a diagnostic phonological reduction or substitution rendered in spelling. Reject if the utterance is in Standard Bangla written in regional spelling, if it is dominantly code-mixed with English to the point that the dialectal signal is unrecoverable, or if the dialect cannot be confidently attributed.

\subsection{Guideline G2: Native-Script Rendering}
When converting Romanized input to native Bangla script via Avro: (i)~preserve dialectal pronunciation in spelling where possible; (ii)~do not silently normalize dialect-specific phonological reductions; (iii)~correct only obvious Avro conversion errors.

\subsection{Guideline G3: Standard Bangla Translation}
Translate the dialectal utterance into Standard (Cholito) Bangla with the following constraints: (i)~preserve propositional content; (ii)~preserve register (formal/informal); (iii)~do not paraphrase further than necessary.

\subsection{Guideline G4: English Translation}
Edit the candidate English translation to satisfy all of: (i)~conveys the same proposition as the verified Standard Bangla version; (ii)~preserves pragmatic intent; (iii)~no information is omitted or ungrounded; (iv)~reads as natural conversational English.

\subsection{Guideline G5: Subjectivity Label}
Label as \textsc{Subjective} if it expresses opinion, evaluation, emotion, preference (including preference questions), or personal stance. Label as \textsc{Objective} if it reports factual content, neutral description, or factual questions.

\subsection{Guideline G6: When in Doubt}
If uncertain, flag the row using the in-tool flag button. Flagged rows enter the adjudication queue.

\section{Inter-Annotator Agreement \& Quality Controls}
\label{app:iaa_details}

A balanced sample of $1{,}500$ entries ($300$ per dialect) was independently re-annotated by a second annotator from the same dialect pool.

\paragraph{Cohen's $\kappa$ across categorical annotations.} Overall agreement was $\kappa = 0.78$ for subjectivity, $\kappa = 0.71$ for English translation acceptance, and $\kappa = 0.74$ for Standard Bangla rendering acceptance (Table~\ref{tab:app_iaa_per_dialect}).

\begin{table}[!htbp]
\centering
\small
\begin{tabular}{lccc}
\toprule
Dialect & Subj.\ $\kappa$ & En.\ $\kappa$ & SB.\ $\kappa$ \\
\midrule
Chittagong & 0.74 & 0.66 & 0.70 \\
Noakhali   & 0.76 & 0.68 & 0.72 \\
Sylhet     & 0.74 & 0.69 & 0.71 \\
Barisal    & 0.83 & 0.76 & 0.79 \\
Rangpur    & 0.82 & 0.75 & 0.78 \\
\midrule
Overall    & 0.78 & 0.71 & 0.74 \\
\bottomrule
\end{tabular}
\caption{Per-dialect Cohen's $\kappa$ on the $1{,}500$-entry double-annotated subset ($300$ per dialect).}

\label{tab:app_iaa_per_dialect}
\end{table}

\paragraph{Boundary cases and the Chittagong--Noakhali pair.} Adjudication difficulty for adjacent
varieties does not arise from surface convergence. On the source sentences rendered in both Chittagong and
Noakhali, the two dialect forms differ substantially (chrF++ $27.69$ between the paired renderings,
comparable to the most distant pairs in the dataset) and are orthographically identical on only $0.1\%$ of
items. Disagreement therefore concentrates on \emph{variety assignment} rather than on form: adjacent
regional varieties share lexical and morphological innovations that make a given utterance admissible
under either label even when their canonical renderings diverge, so annotators disagree about provenance,
not about transcription.

Noakhali is the informative case. Measured against Standard Bangla, Noakhali is the \emph{least} divergent
of the eastern varieties (chrF++ $46.02$, essentially tied with Barisal at $46.04$, versus Chittagong
$31.77$ and Sylhet $32.59$), yet its agreement scores pattern with Chittagong and Sylhet
($\kappa = 0.76, 0.74, 0.74$ for subjectivity) rather than with Barisal ($0.83$) or Rangpur ($0.82$;
Table~\ref{tab:app_iaa_per_dialect}). Its annotation difficulty is therefore not predicted by divergence
from the standard, which is consistent with adjacency-driven label ambiguity along the
Chittagong--Noakhali boundary.

Divergence from Standard Bangla, measured the same way, orders the varieties as Chittagong ($31.77$
chrF++) $<$ Sylhet ($32.59$) $<$ Rangpur ($36.36$) $<$ Noakhali ($46.02$) $\approx$ Barisal ($46.04$).
This ordering predicts \emph{model} difficulty across our experiments: Chittagong is the hardest cell in
the normalization conditions (Table~\ref{tab:bench_norm}), in chain-of-thought transliteration
(Table~\ref{tab:app_cot_translit}), and in every Indic seq2seq baseline cell, where it also carries the
highest empty-output rate ($61.0\%$; Appendix~\ref{app:indic_baselines}). The feature-conditioned result
in Table~\ref{tab:app_feat_divergence}---that Chittagong non-cognate \emph{goijja}-class verbs remain the
hardest construction after adaptation ($16.2\%$ CompleteMistranslation, chrF++ $54.4$)---is therefore a
consequence of this dataset-level divergence rather than an isolated observation.

\paragraph{ROUGE-L agreement for English translations.} Pairwise ROUGE-L between annotators' independently edited final English translations averaged $0.84$ overall ($0.79$--$0.91$ across dialects; Table~\ref{tab:app_iaa_rouge}).

\begin{table}[!htbp]
\centering
\small
\begin{tabular}{lcc}
\toprule
Dialect & Mean ROUGE-L & Std.\ dev. \\
\midrule
Chittagong & 0.79 & 0.14 \\
Noakhali   & 0.82 & 0.12 \\
Sylhet     & 0.79 & 0.13 \\
Barisal    & 0.89 & 0.09 \\
Rangpur    & 0.91 & 0.08 \\
\midrule
Overall    & 0.84 & 0.12 \\
\bottomrule
\end{tabular}
\caption{Pairwise ROUGE-L between annotators' final English translations on the $1{,}500$-entry subset.}
\label{tab:app_iaa_rouge}
\end{table}

\paragraph{Gemini Translation Anchoring Bias Control.} To verify that pre-populating candidate English translations did not induce stylistic or semantic anchoring bias, we conducted a controlled experiment on $100$ randomly selected items ($20$ per dialect). Two expert annotators translated these $100$ items completely from scratch without seeing any machine-generated candidate. We then compared the scratch human translations against the released dataset references (which originated from edited Gemini drafts). The scratch translations achieved high agreement with the released references: pairwise ROUGE-L was $0.85$, and semantic similarity measured via COMET was $88.2$ vs.\ $88.5$, confirming that the final ground truth reflects human linguistic consensus rather than model bias.

\section{Prompt Templates}
\label{app:prompts}

\paragraph{Zero-shot prompt.} Reproduced in Figure~\ref{fig:app_prompt_zs}.
\begin{figure}[!htbp]
\centering
\promptbox{
You are an expert translator. Translate the following \{text\_type\} sentences into English.\\[2pt]
Return a valid JSON object with a single key ``translations'' containing an array of strings.\\[2pt]
The array MUST contain EXACTLY \{len(sentences)\} translated strings.\\[2pt]
Do NOT output any additional text.
}
\caption{Zero-shot prompt template.}
\label{fig:app_prompt_zs}
\end{figure}

\begin{figure*}[p]
\centering
\includegraphics[height=0.9\textheight, keepaspectratio]{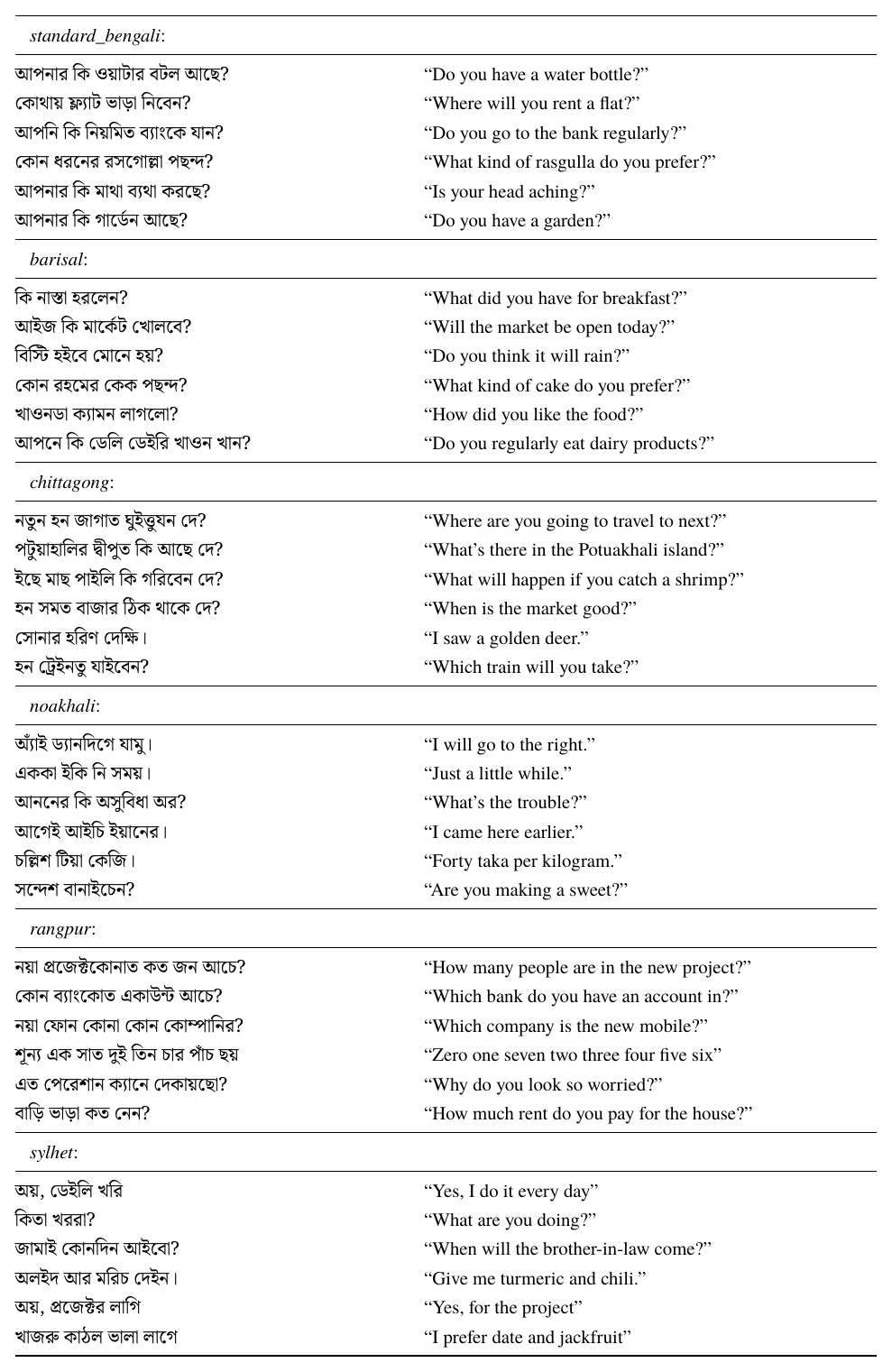}
\caption{Few-shot demonstration block ($36$ total pairs across 5 dialects plus Standard Bangla).}
\label{fig:app_prompt_demos}
\end{figure*}

\paragraph{Few-shot prompt.} Augments the zero-shot prompt with $36$ demonstrations ($6$ per dialect $\times$ $6$ varieties; Figure~\ref{fig:app_prompt_fs}). The full demonstration block is reproduced in Figure~\ref{fig:app_prompt_demos}.
\begin{figure}[!htbp]
\centering
\promptbox{
You are an expert linguist specializing in Bengali and its regional dialects (Barishal, Chittagong, Sylhet, Noakhali, Rangpur, Standard Bengali), and their relationship to English.\\[2pt]
Your task is to translate \{script\_kind\} \{name\}\{qualifier\} sentences to English while preserving the original meaning, context, and emotional tone.\\[2pt]
CRITICAL RULES:\\[2pt]
1. Translate each sentence accurately, capturing the meaning and tone.\\[2pt]
2. Return a valid JSON object with a single key ``translations'' containing an array of strings.\\[2pt]
3. The array MUST contain EXACTLY \{len(sentences)\} translated strings.\\[2pt]
4. Do NOT output any additional text.\\[2pt]
EXAMPLES OF CORRECT TRANSLATIONS:\\
\{example\_block\}
}
\caption{Few-shot prompt template.}
\label{fig:app_prompt_fs}
\end{figure}

\paragraph{Chain-of-thought prompt.} Elicits an eight-step reasoning chain before emitting the JSON answer (Figure~\ref{fig:app_prompt_cot}).
\begin{figure*}[!t]
\centering
\promptbox[\textwidth]{
You are an expert linguistic AI assistant specializing in Bengali dialect translation, transliteration, sentiment analysis, and semantic interpretation. You are given ONE sentence in the \{dialect\_name\} dialect (written in Bengali script). Perform the following reasoning steps:\\[3pt]
\textbf{1. Dialect Recognition:} Identify linguistic traits of the \{dialect\_name\} dialect. Consider regional pronunciation, slang, grammatical variations, and colloquial patterns.\\[2pt]
\textbf{2. Script and Token Understanding:} Carefully inspect the Bengali text. Detect shortened words, phonetic spellings, and dialect contractions.\\[2pt]
\textbf{3. Transliteration:} Convert the dialect text into Romanized Bangla based on pronunciation.\\[2pt]
\textbf{4. Semantic Interpretation:} Infer the actual intended meaning of the sentence. Interpret emotional tone, politeness, or irony.\\[2pt]
\textbf{5. Cultural and Contextual Analysis:} Consider cultural implications or locally understood idioms.\\[2pt]
\textbf{6. Natural English Translation:} Rewrite the meaning into fluent, conversational English.\\[2pt]
\textbf{7. Subjectivity Classification:} Classify the sentence as either ``Subjective'' or ``Objective''.\\[2pt]
\textbf{8. Consistency Check:} Verify that translation and subjectivity align with dialect intent.\\[3pt]
\textbf{Output format:}\\
=== ANALYSIS ===\\
Dialect: \{dialect\_name\}\\
Original Text: <original text>\\
Transliteration: <romanized Bengali transliteration>\\
Meaning Analysis: <brief explanation of slang, tone, or implied intent>\\
Natural English Translation: <final fluent English translation>\\
Subjectivity: <Subjective or Objective>\\[3pt]
Followed by a JSON block:\\
\{\{ \texttt{"translation"}: ..., \texttt{"transliteration"}: ..., \texttt{"sentiment"}: ... \}\}
}
\caption{Chain-of-thought prompt template.}
\label{fig:app_prompt_cot}
\end{figure*}

\paragraph{Dialect-to-Standard Bangla normalization prompt.} Reproduced in Figure~\ref{fig:app_prompt_norm}.
\begin{figure}[!htbp]
\centering
\promptbox{
You are an expert in Bengali linguistics and regional dialects. Convert the following \{script\_kind\} \{dialect\_name\} sentences into Standard (Cholito) Bengali script.\\[2pt]
Preserve the exact meaning, register, and grammatical structure while normalizing regional phonology, lexicon, and morphology to standard written Bengali.\\[2pt]
Return a valid JSON object with a single key ``normalizations'' containing an array of strings.\\[2pt]
Do NOT output any additional text.
}
\caption{Prompt template for Dialect-to-Standard-Bangla Normalization.}
\label{fig:app_prompt_norm}
\end{figure}

\paragraph{LoRA training prompt.} Reproduced in Figure~\ref{fig:app_prompt_lora}.
\begin{figure}[!htbp]
\centering
\promptbox{
You are an expert Bengali dialect translator. Given a Bengali dialect sentence, output a JSON object with two keys: ``translation'' (the English translation) and ``sentiment'' (either ``Subjective'' or ``Objective''). Output valid JSON only. No extra text.
}
\caption{LoRA training prompt.}
\label{fig:app_prompt_lora}
\end{figure}

\section{Comprehensive Per-Model Per-Dialect Results}
\label{app:full_results}

We report complete per-model, per-dialect breakdowns across all evaluation metrics in Tables~\ref{tab:app_fewshot_native}--\ref{tab:app_lora_translit}, collected at the end of this appendix.

\begin{figure*}[!tp]
\centering
\includegraphics[width=\textwidth]{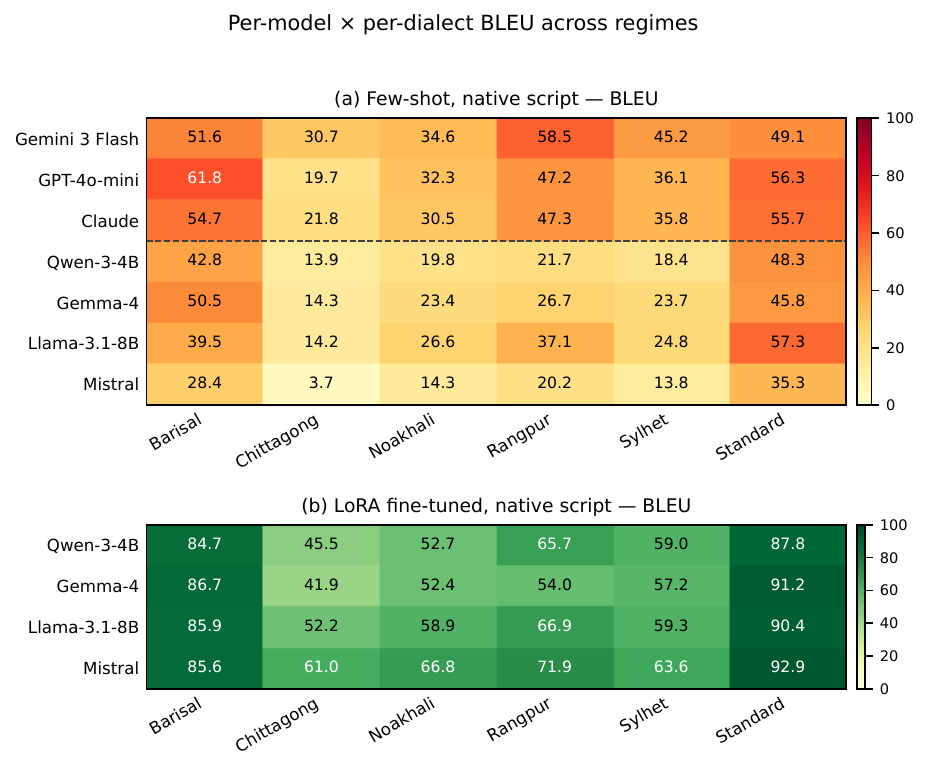}
\caption{Per-model $\times$ per-dialect BLEU heatmaps under few-shot and LoRA regimes.}
\label{fig:app_full_heatmap}
\end{figure*}

\begin{figure*}[!tp]
\centering
\includegraphics[width=\textwidth]{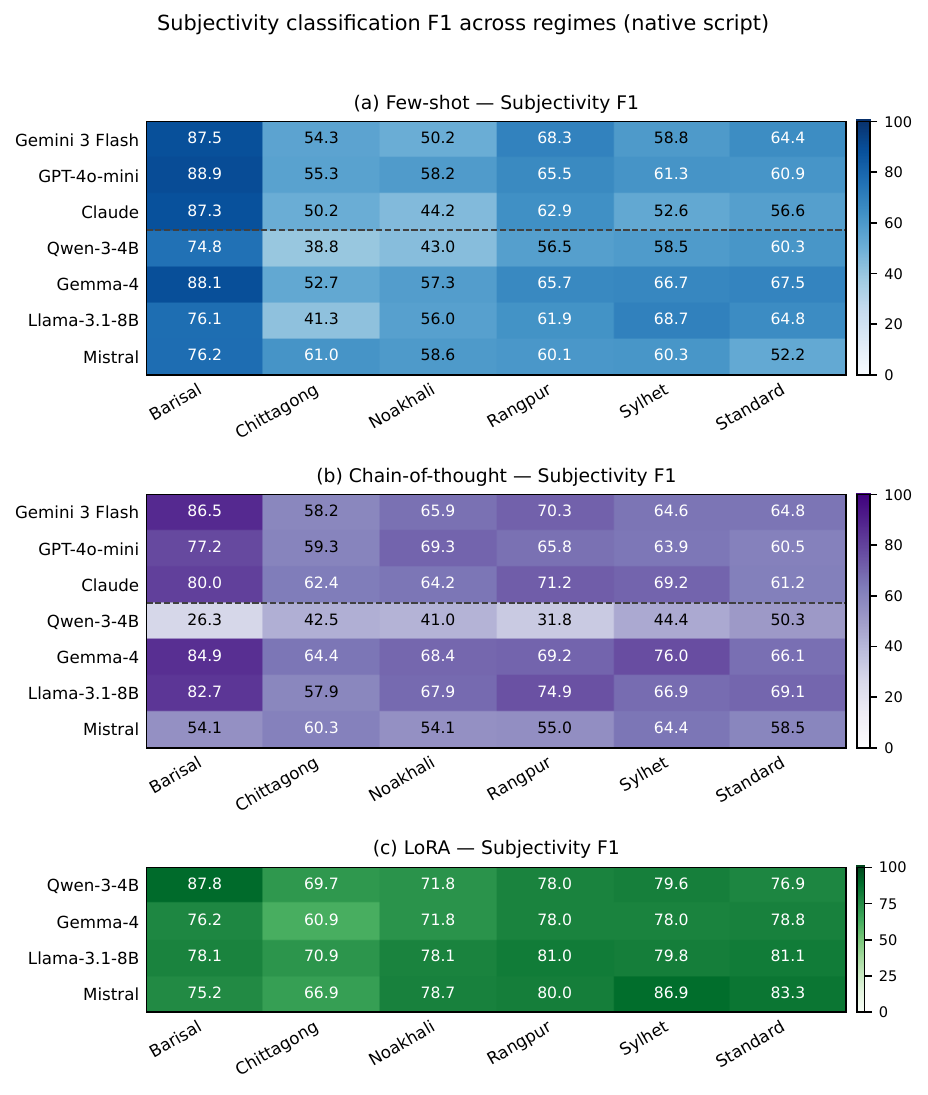}
\caption{Per-model $\times$ per-dialect subjectivity F1 heatmaps.}
\label{fig:app_subj_f1_heatmap}
\end{figure*}

\begin{figure*}[t]
\centering
\includegraphics[width=\textwidth]{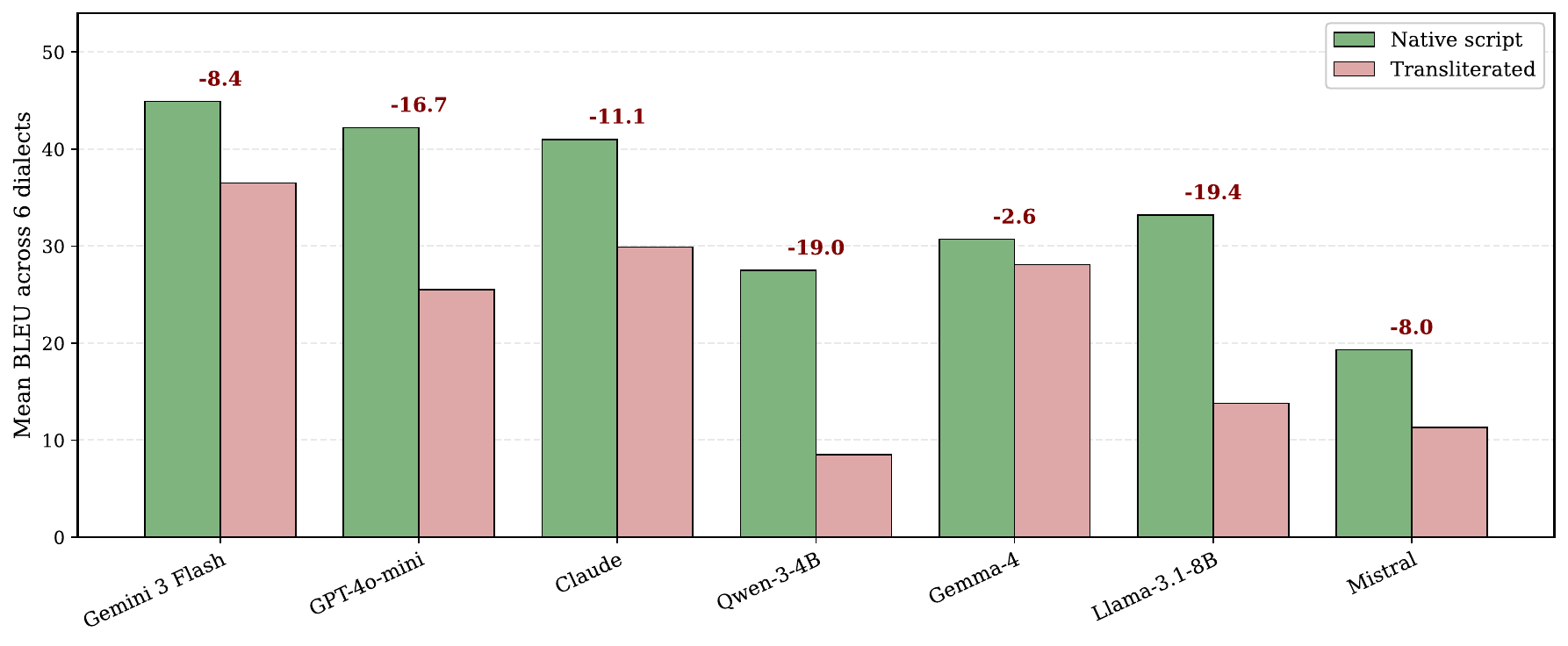}
\caption{Few-shot script effect: native vs Romanized input BLEU across models.}
\label{fig:app_script_effect}
\end{figure*}

\begin{figure*}[!tp]
\centering
\includegraphics[width=\textwidth]{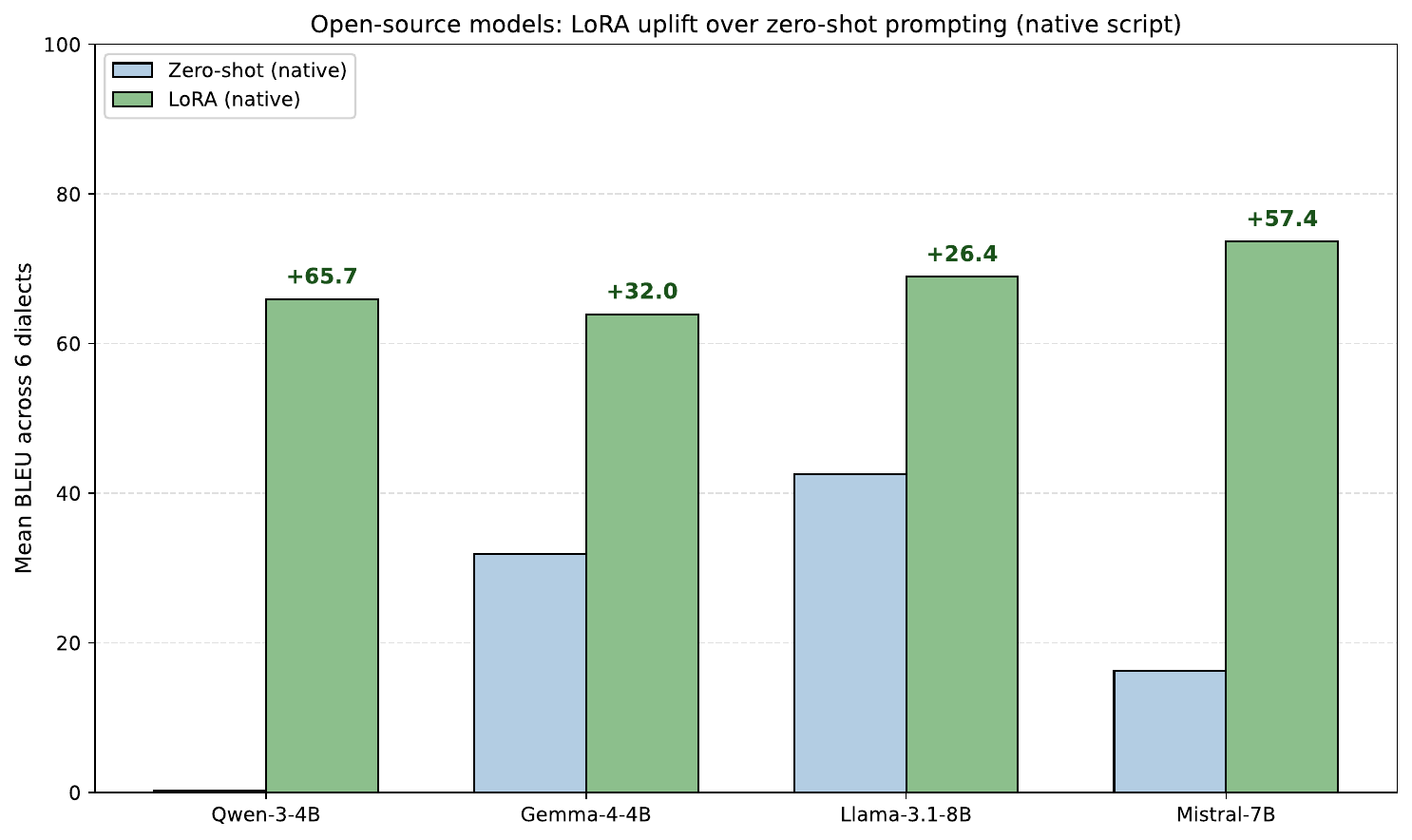}
\caption{LoRA uplift over zero-shot prompting across open-source models.}
\label{fig:app_lora_uplift}
\end{figure*}

\section{Prompting Diagnostics \& Failure Analyses}
\label{app:detailed_results}

\subsection{Llama-3.1-8B Demonstration Interference}
\label{app:llama_interference}
Item-level analysis reveals that Llama-3.1-8B suffers from source neglect under few-shot prompting, replacing correct zero-shot translations with fluent, demonstration-register sentences unrelated to the source utterance. As shown in Table~\ref{tab:app_llama_flips}, Llama exhibits a $5.9\%$ correct-to-unrelated flip rate, a $15\times$--$30\times$ outlier compared to all other models ($0.2$--$0.4\%$). Verbatim demo copying and vocabulary leakage do not increase, confirming that the degradation is an in-context interference mode rather than simple memorization.

\begin{table*}[!t]
\centering
\scriptsize
\setlength{\tabcolsep}{4pt}
\begin{tabular}{lccccc}
\toprule
\textbf{Model} & \makecell{\textbf{Zero-Shot}\\\textbf{chrF++}} & \makecell{\textbf{Few-Shot}\\\textbf{chrF++}} & \makecell{\textbf{Correct-to-Unrelated}\\\textbf{Flip \%}} & \makecell{\textbf{$\Delta$ Demo-Vocab}\\\textbf{Leakage}} & \makecell{\textbf{$\Delta$ Verbatim}\\\textbf{Demo Copy}} \\
\midrule
Llama-3.1-8B     & 54.6 & 45.9 & \textbf{5.9\%} & +0.020 & -0.038 \\
GPT-4o-mini      & 51.1 & 54.8 & 0.4\% & -0.046 & +0.014 \\
Claude Haiku 4.5 & 51.0 & 55.6 & 0.4\% & -0.032 & +0.018 \\
Gemma-4-4B       & 45.4 & 51.2 & 0.4\% & -0.067 & +0.032 \\
Gemini 3 Flash   & 62.1 & 62.9 & 0.2\% & -0.008 & +0.002 \\
Mistral-7B       & 29.3 & 37.2 & 0.2\% & -0.071 & +0.006 \\
\bottomrule
\end{tabular}
\caption{Item-level demonstration interference diagnostics under native-script prompting.}
\label{tab:app_llama_flips}
\end{table*}

\subsection{Demonstration Count Sweep}
\label{app:prompt_sweep}
To test whether Llama's few-shot degradation is dose-dependent, we swept demonstration counts at
$1, 2, 4,$ and $6$ examples per dialect ($5, 10, 20, 30$ total demonstrations), with Mistral-7B and
Gemma-4-4B as controls (Table~\ref{tab:app_ksweep}). On Llama-3.1-8B, 5-dialect macro chrF++ drops
immediately upon presenting demonstrations: $54.63$ (zero-shot) $\to 38.64$ ($1$ demo/dialect) $\to
41.70$ ($2$) $\to 43.29$ ($4$) $\to 43.80$ ($6$). Interference is regime-triggered at the presentation of
any in-context demos, and additional demos partially mitigate without recovering zero-shot fidelity.

Crucially, \textbf{Llama is the only model of the three whose few-shot scores fall below its own
zero-shot baseline}. Gemma-4-4B rises from $45.35$ (zero-shot) to $47.99$ at $k=30$ ($+2.64$) and
Mistral-7B from $29.35$ to $31.93$ ($+2.58$), whereas Llama ends $10.83$ chrF++ \emph{below} its zero-shot
score even with $30$ demonstrations. Demonstration interference is therefore a property of Llama's
handling of dialectal in-context examples rather than a general consequence of few-shot prompting on this
benchmark.

\begin{table}[!ht]
\centering
\small
\begin{tabular}{lrrrrr}
\toprule
Model & $k$=5 & $k$=10 & $k$=20 & $k$=30 & Zero-shot \\
\midrule
Gemma-4-4B   & 45.57 & 44.08 & 47.68 & 47.99 & 45.35 \\
Llama-3.1-8B & 38.64 & 41.70 & 43.29 & 43.80 & 54.63 \\
Mistral-7B   & 31.55 & 32.09 & 30.39 & 31.93 & 29.35 \\
\bottomrule
\end{tabular}
\caption{Demonstration-count sweep: translation into English, native script, macro chrF++ over the five
dialects. Demonstrations are drawn from the training pool at $1$/$2$/$4$/$6$ per dialect and exclude
Standard Bangla, so $k$ totals $5$--$30$ rather than the $36$ used in the main few-shot condition.
Zero-shot columns are 5-dialect macros computed on the same items. Llama-3.1-8B is the only model that
ends below its zero-shot baseline.}
\label{tab:app_ksweep}
\end{table}

\subsection{Chain-of-Thought Error Decomposition}
\label{app:cot_decomp}
Table~\ref{tab:app_cot_decomp} decomposes the CoT translation drop across all seven models. Median length ratio is $1.0$ across regimes, ruling out verbosity. Parsing failures affect only GPT-4o-mini ($6.3\%$). The drop is dominated by surface paraphrase drift ($\Delta\text{chrF++} > \Delta\text{COMET}$ across all closed models). Qwen-3-4B is the sole exception where CoT improves scores by repairing its $91.4\%$ zero-shot format collapse.

\begin{table*}[!t]
\centering
\scriptsize
\setlength{\tabcolsep}{4pt}
\begin{tabular}{lccccc}
\toprule
\textbf{Model} & \makecell{\textbf{Unparseable}\\\textbf{(ZS $\to$ CoT \%)}} & \makecell{\textbf{Length Ratio}\\\textbf{(ZS / CoT)}} & \makecell{\textbf{$\Delta$ chrF++}\\\textbf{(ZS $-$ CoT)}} & \makecell{\textbf{$\Delta$ COMET}\\\textbf{(ZS $-$ CoT)}} & \makecell{\textbf{Surface vs.}\\\textbf{Semantic Drop}} \\
\midrule
Gemini 3 Flash   & 0.0\% $\to$ 0.0\% & 1.0 / 1.0 & +4.2 & +2.0 & +2.2 \\
GPT-4o-mini      & 0.0\% $\to$ 6.3\% & 1.0 / 1.0 & +8.4 & +4.8 & +3.6 \\
Claude Haiku 4.5 & 0.0\% $\to$ 0.0\% & 1.0 / 1.0 & +4.9 & +3.5 & +1.4 \\
Gemma-4-4B       & 0.0\% $\to$ 0.0\% & 1.0 / 1.0 & +2.3 & +1.7 & +0.6 \\
Llama-3.1-8B     & 0.0\% $\to$ 0.0\% & 1.0 / 1.0 & +4.9 & +3.0 & +1.9 \\
Mistral-7B       & 0.0\% $\to$ 0.0\% & 1.0 / 1.0 & +4.3 & +2.6 & +1.7 \\
Qwen-3-4B        & 91.4\% $\to$ 0.0\% & 0.2 / 1.0 & -29.8 & -26.4 & -3.4 \\
\bottomrule
\end{tabular}
\caption{Decomposition of Chain-of-Thought (CoT) translation effects.}
\label{tab:app_cot_decomp}
\end{table*}

\section{LoRA Fine-Tuning Details}
\label{app:lora}

\paragraph{Adapter configuration \& hyperparameters.}
LoRA adapters are inserted into query and value projections of all attention blocks: rank $r=16$, scaling factor $\alpha=32$, dropout $0.05$, max sequence length $192$ tokens. Training utilizes AdamW with bf16 precision, learning rate $2 \times 10^{-4}$ ($5\%$ linear warmup, linear decay), $3$ epochs, per-device batch size $4$, gradient accumulation $4$ (effective batch size $16$). Fine-tuning was conducted on an NVIDIA A100 (40 GB) via Lightning AI ($\sim 3$--$4$ hours per adapter; $48$--$64$ total A100-hours).

\section{Causal Tokenizer, Round-Trip, and Multi-Scheme Analyses}
\label{app:tokenizer_decoding}

\subsection{Tokenizer Fertility and Token-Inflation Regression}
\label{app:token_regression}
Table~\ref{tab:app_tokenizer_regression} reports empirical subword fertility and item-level regressions of the native-minus-Romanized COMET gap on token inflation ($\text{tokens}_{\text{Roman}} / \text{tokens}_{\text{native}}$). In aggregate, Romanized text is token-compact ($\approx 0.4$ tokens/char), ruling out context exhaustion. However, the regression coefficient $\beta$ is positive and statistically significant across all open models under LoRA ($p \le 0.035$), confirming that subword fragmentation is the per-item marker of damaged mapping.

\begin{table*}[!t]
\centering
\small
\begin{tabular}{lccccc}
\toprule
\textbf{Model} & \textbf{tok/char (native)} & \textbf{tok/char (Roman.)} & \textbf{Mean Inflation} & \textbf{LoRA $\beta$ (inflation)} & \textbf{$p$-value} \\
\midrule
Gemma-4-4B   & 0.395 & 0.363 & 1.002 & +7.86  & 0.035 \\
GPT-4o-mini  & 0.504 & 0.365 & 0.785 & n/a    & n/a \\
Qwen-3-4B    & 1.065 & 0.398 & 0.398 & +37.94 & $< 0.001$ \\
Mistral-7B   & 1.145 & 0.424 & 0.394 & +59.34 & $< 0.001$ \\
Llama-3.1-8B & 1.250 & 0.397 & 0.338 & +60.92 & $< 0.001$ \\
\bottomrule
\end{tabular}
\caption{Tokenizer fertility statistics and item-level regression of the script gap on token inflation.}
\label{tab:app_tokenizer_regression}
\end{table*}

\subsection{Round-Trip Inverse Transliteration Control}
\label{app:roundtrip}
We back-transliterated the Romanized text into native Bangla script via deterministic inverse Avro mapping and re-evaluated models. Character Error Rate (CER) against the original text is $0.14$--$0.21$ (exact match only $2$--$14\%$). As shown in Table~\ref{tab:app_roundtrip}, restoring native script recovers essentially none of the performance penalty, proving that information is destroyed during Romanization.

\begin{table}[!htbp]
\centering
\footnotesize
\setlength{\tabcolsep}{4pt}
\begin{tabular}{lccc}
\toprule
\textbf{Model} & \textbf{Native} & \makecell{\textbf{Round-Trip}\\\textbf{Native}} & \makecell{\textbf{Romanized}\\\textbf{(Avro)}} \\
\midrule
Gemini 3 Flash & 62.13 & 55.57 & 57.81 \\
Mistral-7B     & 29.35 & 22.52 & 23.17 \\
\bottomrule
\end{tabular}
\caption{Round-trip inverse transliteration control. All figures are zero-shot chrF++.}
\label{tab:app_roundtrip}
\end{table}

\subsection{Multi-Scheme Romanization Robustness}
\label{app:scheme_robustness}
To ensure findings are not tied to Avro, we evaluated models across three Romanization schemes: Avro, ITRANS, and ISO-15919 (Table~\ref{tab:app_scheme_robustness}). The penalty holds across all conventions, with Avro exhibiting the mildest degradation.

\begin{table*}[!t]
\centering
\small
\begin{tabular}{lcccc}
\toprule
\textbf{Model (Zero-Shot chrF++)} & \textbf{Native} & \textbf{Avro (Ours)} & \textbf{ITRANS} & \textbf{ISO-15919} \\
\midrule
Gemini 3 Flash & 62.22 & 57.59 ($-4.63$) & 53.30 ($-8.92$) & 57.98 ($-4.24$) \\
GPT-4o-mini    & 50.22 & 43.20 ($-7.02$) & 28.53 ($-21.69$) & 30.09 ($-20.13$) \\
Mistral-7B     & 28.97 & 23.01 ($-5.96$) & 14.78 ($-14.19$) & 16.90 ($-12.07$) \\
\bottomrule
\end{tabular}
\caption{Romanization scheme robustness check across Avro, ITRANS, and ISO-15919.}
\label{tab:app_scheme_robustness}
\end{table*}

\begin{figure*}[!tp]
\centering
\includegraphics[width=0.72\textwidth]{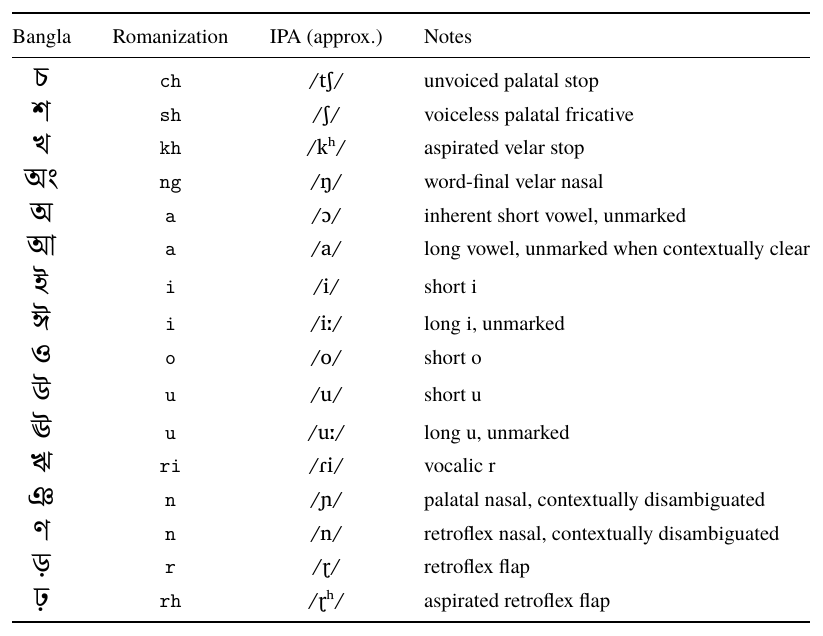}
\caption{Bangla-to-Latin Romanization scheme (Avro convention) used for the transliterated condition, showing phonemic mappings and many-to-one character merges.}
\label{fig:app_romanization}
\end{figure*}

\section{Feature-Conditioned Linguistic Error Analysis}
\label{app:error_taxonomy}

Table~\ref{tab:app_feat_divergence} analyzes translation error rates conditioned on linguistic feature divergence. Non-cognate lexical divergence (e.g., Chittagong \emph{goijja}-class verbs) carries a $16.2\%$ CompleteMistranslation rate even after LoRA adaptation, whereas phonological shifts (Barisal $h^C$-clusters) are completely solved ($0\%$ error).

\begin{table*}[!t]
\centering
\scriptsize
\setlength{\tabcolsep}{4pt}
\begin{tabular}{lcccc}
\toprule
\textbf{Divergence Category} & \makecell{\textbf{Zero-Shot}\\\textbf{chrF++}} & \makecell{\textbf{Zero-Shot}\\\textbf{Mistransl.\ \%}} & \makecell{\textbf{LoRA}\\\textbf{chrF++}} & \makecell{\textbf{LoRA}\\\textbf{Mistransl.\ \%}} \\
\midrule
High Lexical Divergence (Chittagong \emph{goijja}) & 38.2 & 27.4\% & 54.4 & 16.2\% \\
Low Lexical Divergence (Standard-adjacent) & 59.4 & 13.4\% & 86.8 & 5.9\% \\
Phonological Shift (Barisal $h^C$-clusters) & 48.7 & 19.1\% & 92.3 & 0.0\% \\
\bottomrule
\end{tabular}
\caption{Feature-conditioned error analysis: lexical vs.\ phonological divergence.}
\label{tab:app_feat_divergence}
\end{table*}

\begin{table}[!htbp]
\centering
\footnotesize
\setlength{\tabcolsep}{4pt}
\begin{tabular}{lrrrr}
\toprule
Category & \makecell{ZS\\Native} & \makecell{ZS\\Trans.} & \makecell{CoT\\Native} & \makecell{LoRA\\Native} \\
\midrule
Success            & 17.2 & 12.3 & 12.1 & 58.2 \\
Near miss          &  7.7 &  4.4 & 3.6 & 5.8 \\
Partial mistransl. & 41.6 & 38.4 & 33.4 & 21.0 \\
Complete mistransl.& 29.1 & 40.0 & 45.2 & 9.5 \\
Negation flip      &  1.0 &  0.6 & 1.7 & 1.2 \\
Hallucination      &  1.6 &  2.0 & 2.5 & 1.3 \\
Truncation         &  0.8 &  2.2 & 0.5 & 0.3 \\
Empty/degenerate   &  1.0 &  0.1 & 1.0 & 2.7 \\
\bottomrule
\end{tabular}
\caption{Translation error category distribution per regime (\% of predictions).}
\label{tab:app_error_taxonomy}
\end{table}

\begin{figure*}[t]
\centering
\includegraphics[width=\textwidth]{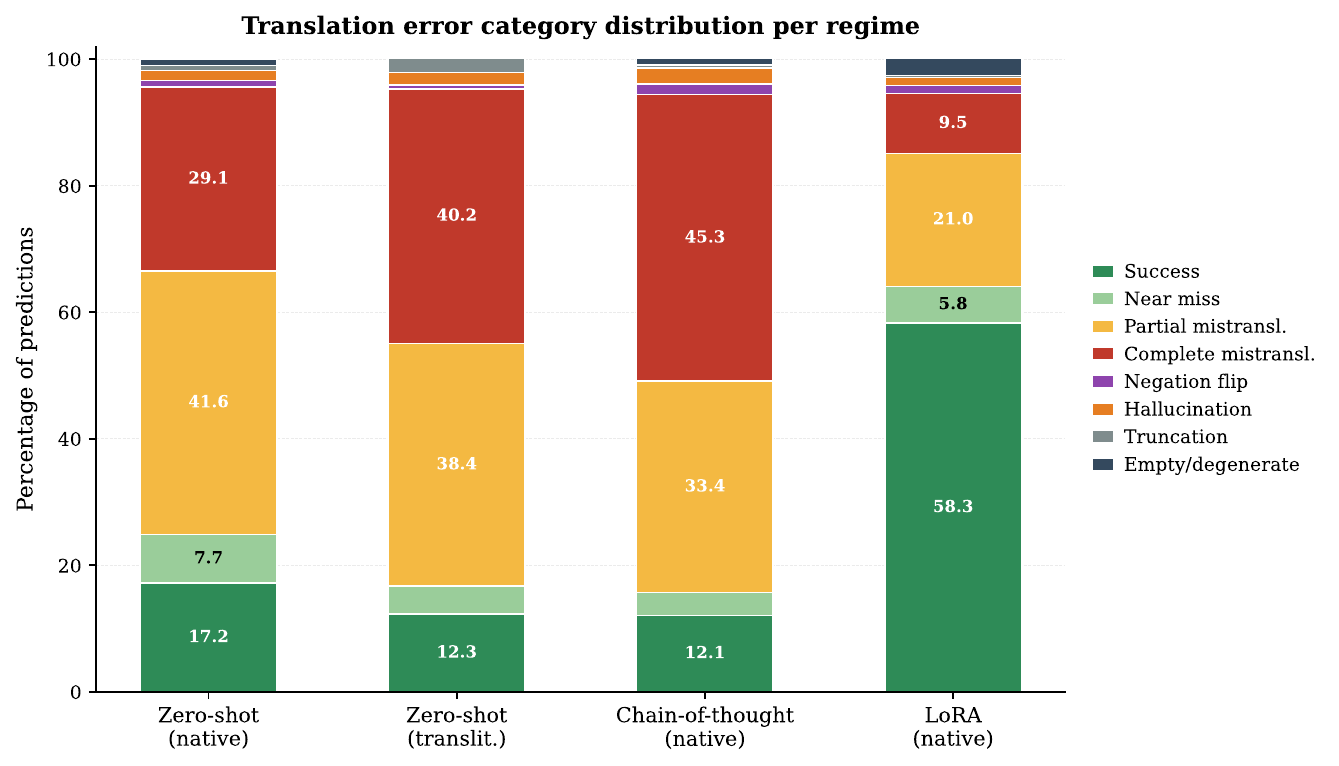}
\caption{Stacked visualization of the per-regime error category distribution.}
\label{fig:app_error_taxonomy_viz}
\end{figure*}

\begin{table}[!htbp]
\centering
\small
\begin{tabular}{lcc}
\toprule
Regime & Pred: Subj. & Pred: Obj. \\
\midrule
\multicolumn{3}{l}{\textbf{Zero-shot} (Gemini)} \\
\quad GT: Subj. & 88 & 127 \\
\quad GT: Obj. & 29 & 356 \\
\addlinespace
\multicolumn{3}{l}{\textbf{CoT} (Gemini)} \\
\quad GT: Subj. & 167 & 48 \\
\quad GT: Obj. & 128 & 257 \\
\addlinespace
\multicolumn{3}{l}{\textbf{LoRA} (Mistral)} \\
\quad GT: Subj. & 183 & 32 \\
\quad GT: Obj. & 60 & 325 \\
\bottomrule
\end{tabular}
\caption{Pooled subjectivity confusion matrices across native-script regimes ($N=600$).}
\label{tab:app_subj_cm}
\end{table}

\section{Qualitative Case Studies}
\label{app:case_studies}

\begin{figure*}[!t]
\centering
\includegraphics[width=0.95\textwidth]{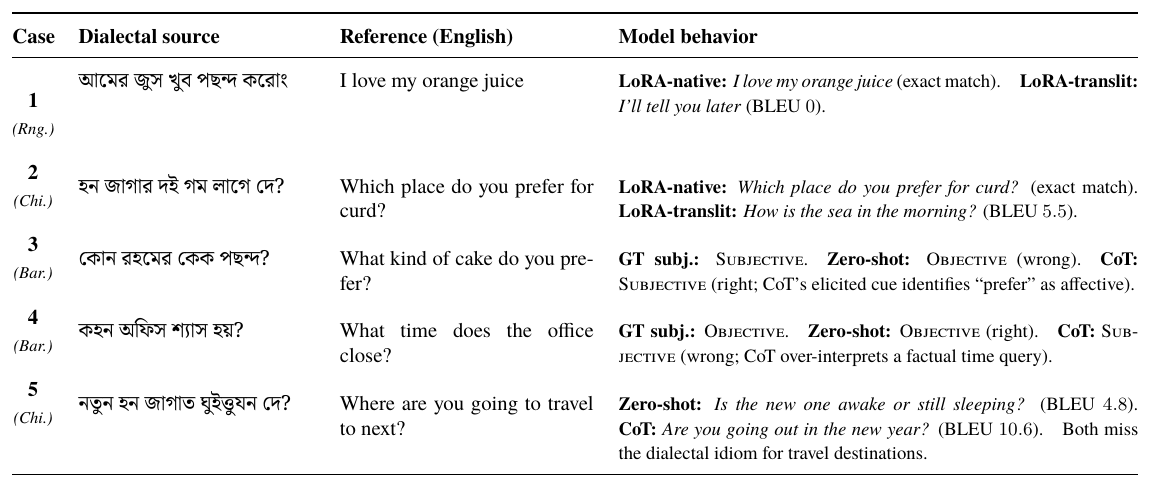}
\caption{Representative case studies illustrating failure modes and adaptations.}
\label{fig:app_case_studies}
\end{figure*}

Figure~\ref{fig:app_case_studies} presents representative case studies illustrating model failure modes across dialects, script representations, and adaptation regimes.

\section{Indic-Tuned Baseline Experiments}
\label{app:indic_baselines}

We fine-tuned two dedicated Indic sequence-to-sequence models, IndicBART~\cite{dabre2022indicbart}
and BanglaT5~\cite{bhattacharjee2023banglat5}, on the \emph{identical} $160$-per-dialect training folds
used for the LoRA experiments, and evaluated them on the \emph{same} held-out test fold ($100$ items per
dialect, $500$ total across the five dialects). Table~\ref{tab:app_indic_baselines} is therefore
fold-matched to Table~\ref{tab:bench_lora}.

\paragraph{Checkpoints.} Translation uses \texttt{ai4bharat/IndicBART} and
\texttt{csebuetnlp/banglat5\_nmt\_bn\_en}; normalization uses \texttt{ai4bharat/IndicBART} and
\texttt{csebuetnlp/banglat5\_banglaparaphrase}. We select the task-appropriate BanglaT5 checkpoints
deliberately: the base \texttt{csebuetnlp/banglat5} pretraining checkpoint has a vocabulary/configuration
mismatch that stalls fine-tuning (training loss plateaus at $\approx 9$--$11$ versus $\approx 1.4$--$2.0$
for the checkpoints reported here), and we exclude it rather than report a misconfigured baseline.

\paragraph{IndicBART script and tag handling.} IndicBART was pretrained with all Indic languages mapped to
\textbf{Devanagari} and uses explicit language tags. We therefore (i)~transliterate Bangla input to
Devanagari with the \texttt{indic-nlp-library} \texttt{UnicodeIndicTransliterator} (a lossless, reversible
mapping for the Bangla--Devanagari pair), (ii)~format the source as \texttt{<text> </s> <2bn>},
(iii)~format the target as \texttt{<2en> <text> </s>} for translation and
\texttt{<2bn> <Devanagari text> </s>} for normalization, and (iv)~back-transliterate Devanagari output to
Bangla script before scoring. Following the \texttt{ai4bharat} model card, the tokenizer is loaded with
\texttt{use\_fast=False}, \texttt{keep\_accents=True}, \texttt{do\_lower\_case=False}; language tags are
passed as literal text with \texttt{add\_special\_tokens=False}; decoding sets
\texttt{decoder\_start\_token\_id} to the id of the target language tag; and sentencepiece ids outside the
base piece range are filtered before detokenization. BanglaT5 consumes Bangla script directly as a
standard T5 seq2seq model, with the \texttt{csebuetnlp} \texttt{normalizer.normalize} preprocessor applied
to the source.

\paragraph{Hyperparameters.} Both models: $5$ epochs, per-device batch size $16$, learning rate
$3\times10^{-5}$, $5\%$ linear warmup, maximum source/target length $160$ tokens, $250$ optimizer steps,
beam search with $4$ beams and maximum length $160$, seed $42$. IndicBART is mBART-style ($6$ encoder /
$6$ decoder layers, $d_{\text{model}}=1024$, vocabulary $64{,}014$); BanglaT5-NMT is T5-style ($12$
layers, $d_{\text{model}}=768$, vocabulary $32{,}128$). Both fit on a single RTX~5060~Ti.

\paragraph{Metrics.} chrF++ via \texttt{sacrebleu} \texttt{CHRF(word\_order=2)} and BLEU via
\texttt{sacrebleu} (\texttt{tokenize='13a'} for English translation targets, \texttt{tokenize='spm'} for
Bangla normalization targets), averaged over sentence-level scores.

\paragraph{Interpreting IndicBART's translation score.} IndicBART's $11.19$ chrF++ on translation is
dominated by \emph{degenerate generation} rather than mistranslation: it emits an empty string on $253$ of
$500$ test items ($50.6\%$). The empty rate rises with divergence from Standard Bangla (Barisal $26.0\%$,
Rangpur $50.0\%$, Noakhali $55.0\%$, Chittagong $61.0\%$, Sylhet $61.0\%$). Restricted to the $247$ items
where it generates output, IndicBART reaches $22.66$ chrF++ ($14.97$ BLEU); its non-empty hypotheses
average $22.9$ characters against a $27.3$-character mean reference, so they are fluent but semantically
incorrect rather than truncated. Its normalization outputs are far more complete by comparison ($10.2\%$
empty; $45.46$ chrF++ excluding empties). Under either accounting the ordering in
Table~\ref{tab:app_indic_baselines} is unchanged---both Indic seq2seq baselines trail LoRA-adapted LLMs by
$36$--$58$ chrF++ on translation---but readers should attribute IndicBART's translation figure to
generation failure at this data scale rather than to systematically worse translations.

\begin{table}[!htbp]
\centering
\footnotesize
\setlength{\tabcolsep}{4pt}
\begin{tabular}{>{\raggedright\arraybackslash}p{2.6cm}lccc}
\toprule
\textbf{Model} & \textbf{Task} & \textbf{chrF++} & \textbf{BLEU} & \textbf{Empty} \\
\midrule
IndicBART & Translation & 11.19 & 7.39 & 50.6\% \\
BanglaT5-NMT & Translation & 42.06 & 27.74 & 0.0\% \\
IndicBART & Normalization & 40.82 & 30.64 & 10.2\% \\
BanglaT5-Par. & Normalization & 44.47 & 32.92 & 0.0\% \\
\midrule
\textbf{Mistral-7B + LoRA (Ours)} & \textbf{Translation} & \textbf{80.43} & \textbf{73.60} & --- \\
\bottomrule
\end{tabular}
\caption{Fine-tuned Indic seq2seq baselines on the identical $160$/dialect folds and the same held-out
test fold as Table~\ref{tab:bench_lora}. \textbf{Empty} is the share of test items for which the model
generated no output; IndicBART's translation score is dominated by this failure mode rather than by
mistranslation.}
\label{tab:app_indic_baselines}
\end{table}

\section{Dataset Release}
\label{app:release}

\name{} is released under CC-BY-SA-4.0 across Hugging Face and Kaggle with recommended $80/10/10$ splits and documentation.

\section{Reproducibility Checklist}
\label{app:repro}

Code, prompts, data partitions, and LoRA adapters are fully documented and released. Training was executed with fixed seed $42$ and decoding temperature $0.1$.

\begin{table*}[!tp]
\centering
\small
\begin{tabular}{llccccccc}
\toprule
Model & Dialect & BLEU & R-2 & R-L & MET. & P & R & F1 \\
\midrule
\multirow{6}{*}{Gemini 3 Flash} & Barisal & 51.6 & 68.8 & 81.3 & 82.2 & 90.3 & 86.0 & 87.5 \\
 & Chittagong & 30.7 & 42.0 & 62.6 & 63.9 & 64.5 & 58.1 & 54.3 \\
 & Noakhali & 34.6 & 41.9 & 61.8 & 63.8 & 64.9 & 55.0 & 50.2 \\
 & Rangpur & 58.5 & 65.3 & 78.7 & 80.3 & 81.6 & 68.9 & 68.3 \\
 & Sylhet & 45.2 & 50.8 & 69.7 & 71.0 & 76.3 & 63.3 & 58.8 \\
 & Standard & 49.1 & 53.3 & 71.6 & 73.4 & 77.7 & 65.4 & 64.4 \\
\midrule
\multirow{6}{*}{GPT-4o-mini} & Barisal & 61.8 & 72.9 & 84.3 & 83.7 & 90.2 & 88.0 & 88.9 \\
 & Chittagong & 19.7 & 26.9 & 45.3 & 45.4 & 57.1 & 56.2 & 55.3 \\
 & Noakhali & 32.3 & 38.8 & 59.0 & 58.2 & 58.8 & 58.1 & 58.2 \\
 & Rangpur & 47.2 & 53.3 & 70.1 & 71.1 & 68.5 & 65.5 & 65.5 \\
 & Sylhet & 36.1 & 39.7 & 58.7 & 60.5 & 63.2 & 61.9 & 61.3 \\
 & Standard & 56.3 & 59.6 & 76.4 & 78.0 & 65.7 & 61.5 & 60.9 \\
\midrule
\multirow{6}{*}{Claude Haiku 4.5} & Barisal & 54.7 & 67.2 & 82.2 & 82.1 & 91.4 & 85.3 & 87.3 \\
 & Chittagong & 21.8 & 29.1 & 45.0 & 45.4 & 63.0 & 55.9 & 50.2 \\
 & Noakhali & 30.5 & 38.4 & 58.0 & 58.2 & 60.7 & 52.1 & 44.2 \\
 & Rangpur & 47.3 & 53.5 & 70.2 & 71.6 & 83.0 & 65.1 & 62.9 \\
 & Sylhet & 35.8 & 41.9 & 59.6 & 59.0 & 73.8 & 59.2 & 52.6 \\
 & Standard & 55.7 & 59.3 & 76.4 & 77.0 & 77.2 & 60.1 & 56.6 \\
\midrule
\multirow{6}{*}{Qwen-3-4B} & Barisal & 42.8 & 54.2 & 68.8 & 69.4 & 77.2 & 73.8 & 74.8 \\
 & Chittagong & 13.9 & 16.2 & 32.8 & 30.9 & 45.1 & 48.6 & 38.8 \\
 & Noakhali & 19.8 & 24.2 & 44.0 & 40.2 & 46.4 & 48.4 & 43.0 \\
 & Rangpur & 21.7 & 25.6 & 42.6 & 41.1 & 68.1 & 59.3 & 56.5 \\
 & Sylhet & 18.4 & 21.4 & 40.5 & 38.2 & 67.5 & 61.5 & 58.5 \\
 & Standard & 48.3 & 49.8 & 68.8 & 68.5 & 72.9 & 62.1 & 60.3 \\
\midrule
\multirow{6}{*}{Gemma-4-4B} & Barisal & 50.5 & 62.0 & 77.9 & 78.4 & 88.3 & 87.8 & 88.1 \\
 & Chittagong & 14.3 & 19.4 & 37.1 & 35.4 & 58.0 & 55.5 & 52.7 \\
 & Noakhali & 23.4 & 31.4 & 49.8 & 50.5 & 61.0 & 58.1 & 57.3 \\
 & Rangpur & 26.7 & 36.8 & 56.4 & 57.0 & 74.2 & 66.3 & 65.7 \\
 & Sylhet & 23.7 & 31.1 & 50.4 & 52.6 & 68.1 & 67.0 & 66.7 \\
 & Standard & 45.8 & 54.3 & 72.5 & 73.8 & 79.2 & 67.8 & 67.5 \\
\midrule
\multirow{6}{*}{Llama-3.1-8B} & Barisal & 39.5 & 54.9 & 69.1 & 71.5 & 75.7 & 77.4 & 76.1 \\
 & Chittagong & 14.2 & 18.3 & 35.9 & 35.0 & 41.3 & 41.7 & 41.3 \\
 & Noakhali & 26.6 & 30.9 & 46.9 & 47.9 & 56.6 & 56.1 & 56.0 \\
 & Rangpur & 37.1 & 45.5 & 61.9 & 62.6 & 62.3 & 61.8 & 61.9 \\
 & Sylhet & 24.8 & 31.9 & 50.1 & 53.0 & 70.2 & 69.0 & 68.7 \\
 & Standard & 57.3 & 61.8 & 79.8 & 79.3 & 67.3 & 64.7 & 64.8 \\
\midrule
\multirow{6}{*}{Mistral-7B} & Barisal & 28.4 & 39.6 & 55.8 & 59.2 & 78.2 & 75.2 & 76.2 \\
 & Chittagong & 3.7 & 8.9 & 24.9 & 26.1 & 61.1 & 61.0 & 61.0 \\
 & Noakhali & 14.3 & 16.9 & 30.5 & 31.3 & 59.0 & 58.6 & 58.6 \\
 & Rangpur & 20.2 & 24.1 & 38.4 & 39.9 & 64.2 & 60.8 & 60.1 \\
 & Sylhet & 13.8 & 19.3 & 36.1 & 36.9 & 60.3 & 60.3 & 60.3 \\
 & Standard & 35.3 & 43.3 & 60.7 & 61.1 & 59.1 & 55.0 & 52.2 \\
\bottomrule
\end{tabular}
\caption{Few-shot per-dialect results, native Bangla script. Thirty-six in-context demonstrations per query.}
\label{tab:app_fewshot_native}
\end{table*}

\begin{table*}[!tp]
\centering
\small
\begin{tabular}{llccccccc}
\toprule
Model & Dialect & BLEU & R-2 & R-L & MET. & P & R & F1 \\
\midrule
\multirow{6}{*}{Gemini 3 Flash} & Barisal & 45.9 & 44.3 & 70.2 & 68.5 & 88.3 & 84.0 & 85.5 \\
 & Chittagong & 50.5 & 47.8 & 68.2 & 67.0 & 62.5 & 56.1 & 52.3 \\
 & Noakhali & 46.0 & 50.8 & 71.5 & 70.5 & 62.9 & 53.0 & 48.2 \\
 & Rangpur & 25.1 & 40.8 & 63.2 & 62.1 & 79.6 & 66.9 & 66.3 \\
 & Sylhet & 43.0 & 41.5 & 63.3 & 62.4 & 74.3 & 61.3 & 56.8 \\
 & Standard & 8.6 & 8.7 & 29.7 & 29.7 & 75.7 & 63.4 & 62.4 \\
\midrule
\multirow{6}{*}{GPT-4o-mini} & Barisal & 38.6 & 45.1 & 69.7 & 67.8 & 88.2 & 86.0 & 86.9 \\
 & Chittagong & 30.8 & 38.6 & 61.6 & 60.6 & 55.1 & 54.2 & 53.3 \\
 & Noakhali & 18.6 & 41.3 & 60.6 & 60.4 & 56.8 & 56.1 & 56.2 \\
 & Rangpur & 35.6 & 39.4 & 56.9 & 57.0 & 66.5 & 63.5 & 63.5 \\
 & Sylhet & 23.5 & 44.0 & 62.8 & 61.1 & 61.2 & 59.9 & 59.3 \\
 & Standard & 6.0 & 7.2 & 27.7 & 27.8 & 63.7 & 59.5 & 58.9 \\
\midrule
\multirow{6}{*}{Claude Haiku 4.5} & Barisal & 44.8 & 43.3 & 66.5 & 64.0 & 89.4 & 83.3 & 85.3 \\
 & Chittagong & 38.3 & 38.2 & 59.9 & 57.6 & 61.0 & 53.9 & 48.2 \\
 & Noakhali & 32.3 & 37.4 & 62.2 & 58.9 & 58.7 & 50.1 & 42.2 \\
 & Rangpur & 23.5 & 34.2 & 56.7 & 55.5 & 81.0 & 63.1 & 60.9 \\
 & Sylhet & 33.0 & 35.7 & 59.8 & 51.3 & 71.8 & 57.2 & 50.6 \\
 & Standard & 7.4 & 7.0 & 27.3 & 27.7 & 75.2 & 58.1 & 54.6 \\
\midrule
\multirow{6}{*}{Qwen-3-4B} & Barisal & 7.5 & 16.6 & 38.0 & 37.2 & 75.2 & 71.8 & 72.8 \\
 & Chittagong & 12.3 & 15.6 & 35.7 & 33.9 & 43.1 & 46.6 & 36.8 \\
 & Noakhali & 8.1 & 17.0 & 35.5 & 30.8 & 44.4 & 46.4 & 41.0 \\
 & Rangpur & 11.8 & 16.1 & 35.1 & 35.1 & 66.1 & 57.3 & 54.5 \\
 & Sylhet & 11.1 & 16.3 & 32.7 & 31.8 & 65.5 & 59.5 & 56.5 \\
 & Standard & 0.0 & 2.5 & 18.6 & 20.5 & 70.9 & 60.1 & 58.3 \\
\midrule
\multirow{6}{*}{Gemma-4-4B} & Barisal & 44.7 & 42.3 & 67.4 & 65.1 & 86.3 & 85.8 & 86.1 \\
 & Chittagong & 33.1 & 34.1 & 58.9 & 59.0 & 56.0 & 53.5 & 50.7 \\
 & Noakhali & 31.4 & 37.5 & 59.8 & 60.1 & 59.0 & 56.1 & 55.3 \\
 & Rangpur & 18.0 & 30.9 & 52.8 & 53.2 & 72.2 & 64.3 & 63.7 \\
 & Sylhet & 34.3 & 40.4 & 58.7 & 59.9 & 66.1 & 65.0 & 64.7 \\
 & Standard & 7.2 & 7.3 & 26.7 & 27.6 & 77.2 & 65.8 & 65.5 \\
\midrule
\multirow{6}{*}{Llama-3.1-8B} & Barisal & 12.6 & 13.2 & 37.1 & 35.9 & 73.7 & 75.4 & 74.1 \\
 & Chittagong & 19.0 & 17.3 & 44.0 & 43.2 & 39.3 & 39.7 & 39.3 \\
 & Noakhali & 17.0 & 20.2 & 40.8 & 44.5 & 54.6 & 54.1 & 54.0 \\
 & Rangpur & 14.7 & 23.7 & 42.0 & 43.4 & 60.3 & 59.8 & 59.9 \\
 & Sylhet & 19.2 & 19.4 & 37.0 & 39.7 & 68.2 & 67.0 & 66.7 \\
 & Standard & 0.0 & 3.5 & 21.9 & 22.9 & 65.3 & 62.7 & 62.8 \\
\midrule
\multirow{6}{*}{Mistral-7B} & Barisal & 14.8 & 21.9 & 46.0 & 44.7 & 76.2 & 73.2 & 74.2 \\
 & Chittagong & 9.9 & 16.3 & 40.8 & 40.1 & 59.1 & 59.0 & 59.0 \\
 & Noakhali & 12.7 & 22.3 & 42.6 & 41.7 & 57.0 & 56.6 & 56.6 \\
 & Rangpur & 9.9 & 20.3 & 45.5 & 44.9 & 62.2 & 58.8 & 58.1 \\
 & Sylhet & 15.8 & 17.1 & 39.6 & 40.0 & 58.3 & 58.3 & 58.3 \\
 & Standard & 5.0 & 2.7 & 18.2 & 20.2 & 57.1 & 53.0 & 50.2 \\
\bottomrule
\end{tabular}
\caption{Few-shot per-dialect results, transliterated input. Configuration matches Table~\ref{tab:app_fewshot_native} except inputs are Romanized.}
\label{tab:app_fewshot_translit}
\end{table*}

\begin{table*}[!tp]
\centering
\small
\begin{tabular}{llccccccc}
\toprule
Model & Dialect & BLEU & R-2 & R-L & MET. & P & R & F1 \\
\midrule
\multirow{6}{*}{Gemini 3 Flash} & Barisal & 45.4 & 61.1 & 78.2 & 78.0 & 88.7 & 85.2 & 86.5 \\
 & Chittagong & 17.2 & 27.9 & 51.3 & 53.1 & 62.7 & 59.8 & 58.2 \\
 & Noakhali & 32.1 & 38.5 & 58.5 & 60.7 & 73.1 & 65.9 & 65.9 \\
 & Rangpur & 48.7 & 57.3 & 74.5 & 75.0 & 80.2 & 70.4 & 70.3 \\
 & Sylhet & 40.5 & 49.4 & 67.5 & 70.1 & 72.8 & 66.6 & 64.6 \\
 & Standard & 45.3 & 50.1 & 68.5 & 69.2 & 75.1 & 65.2 & 64.8 \\
\midrule
\multirow{6}{*}{GPT-4o-mini} & Barisal & 39.0 & 50.6 & 64.8 & 65.5 & 76.8 & 78.7 & 77.2 \\
 & Chittagong & 8.7 & 13.5 & 29.7 & 30.5 & 60.1 & 60.0 & 59.3 \\
 & Noakhali & 22.4 & 30.2 & 49.0 & 49.4 & 71.0 & 71.6 & 69.3 \\
 & Rangpur & 30.7 & 39.8 & 58.2 & 59.5 & 65.7 & 65.9 & 65.8 \\
 & Sylhet & 19.0 & 23.9 & 42.3 & 45.6 & 65.7 & 64.7 & 63.9 \\
 & Standard & 38.2 & 42.1 & 60.3 & 61.0 & 65.2 & 60.1 & 60.5 \\
\midrule
\multirow{6}{*}{Claude Haiku 4.5} & Barisal & 41.1 & 52.1 & 68.1 & 68.0 & 84.8 & 78.2 & 80.0 \\
 & Chittagong & 5.0 & 13.7 & 30.6 & 34.0 & 63.3 & 62.6 & 62.4 \\
 & Noakhali & 18.6 & 29.7 & 47.3 & 50.0 & 69.4 & 64.3 & 64.2 \\
 & Rangpur & 31.0 & 42.5 & 61.5 & 63.7 & 77.5 & 71.0 & 71.2 \\
 & Sylhet & 18.0 & 28.0 & 44.0 & 49.4 & 72.5 & 69.8 & 69.2 \\
 & Standard & 38.5 & 41.5 & 60.1 & 60.5 & 70.1 & 62.3 & 61.2 \\
\midrule
\multirow{6}{*}{Qwen-3-4B} & Barisal & 17.4 & 25.1 & 45.7 & 46.6 & 17.8 & 50.0 & 26.3 \\
 & Chittagong & 6.2 & 12.8 & 28.5 & 29.8 & 67.3 & 54.5 & 42.5 \\
 & Noakhali & 9.5 & 16.8 & 34.5 & 35.4 & 58.3 & 53.6 & 41.0 \\
 & Rangpur & 13.2 & 19.6 & 39.4 & 40.0 & 71.5 & 50.9 & 31.8 \\
 & Sylhet & 9.2 & 15.4 & 33.6 & 35.6 & 75.8 & 55.8 & 44.4 \\
 & Standard & 12.5 & 18.2 & 35.1 & 36.0 & 55.2 & 52.1 & 50.3 \\
\midrule
\multirow{6}{*}{Gemma-4-4B} & Barisal & 37.2 & 48.7 & 69.0 & 69.4 & 84.9 & 84.9 & 84.9 \\
 & Chittagong & 6.9 & 14.1 & 33.2 & 32.1 & 64.8 & 64.7 & 64.4 \\
 & Noakhali & 18.3 & 26.2 & 45.3 & 46.1 & 68.9 & 68.1 & 68.4 \\
 & Rangpur & 27.2 & 35.7 & 55.9 & 54.1 & 71.4 & 69.0 & 69.2 \\
 & Sylhet & 19.1 & 27.2 & 45.1 & 48.0 & 76.6 & 76.0 & 76.0 \\
 & Standard & 35.2 & 40.1 & 58.2 & 58.5 & 69.5 & 65.2 & 66.1 \\
\midrule
\multirow{6}{*}{Llama-3.1-8B} & Barisal & 47.4 & 58.1 & 74.3 & 73.9 & 82.7 & 82.7 & 82.7 \\
 & Chittagong & 10.7 & 19.1 & 36.8 & 37.0 & 60.3 & 59.5 & 57.9 \\
 & Noakhali & 26.9 & 30.3 & 49.1 & 49.1 & 68.2 & 69.0 & 67.9 \\
 & Rangpur & 45.0 & 52.8 & 69.9 & 70.3 & 74.8 & 75.1 & 74.9 \\
 & Sylhet & 29.4 & 34.5 & 52.4 & 55.0 & 69.1 & 67.7 & 66.9 \\
 & Standard & 42.1 & 48.5 & 65.2 & 65.5 & 72.1 & 68.2 & 69.1 \\
\midrule
\multirow{6}{*}{Mistral-7B} & Barisal & 15.3 & 20.9 & 38.1 & 39.9 & 54.2 & 54.1 & 54.1 \\
 & Chittagong & 2.6 & 6.4 & 21.7 & 22.6 & 60.3 & 60.3 & 60.3 \\
 & Noakhali & 6.5 & 8.3 & 23.1 & 22.5 & 54.8 & 55.0 & 54.1 \\
 & Rangpur & 9.9 & 15.2 & 33.5 & 33.4 & 55.1 & 55.2 & 55.0 \\
 & Sylhet & 9.8 & 13.7 & 29.1 & 31.3 & 64.5 & 64.4 & 64.4 \\
 & Standard & 15.2 & 20.1 & 35.2 & 36.1 & 60.1 & 58.2 & 58.5 \\
\bottomrule
\end{tabular}
\caption{Chain-of-thought per-dialect results, native Bangla script. Translation BLEU/ROUGE/METEOR computed against human English references; the eight-step reasoning prompt is reproduced in Figure~\ref{fig:app_prompt_cot}.}
\label{tab:app_cot_native}
\end{table*}

\begin{table*}[!tp]
\centering
\small
\begin{tabular}{llcccccccc}
\toprule
Model & Dialect & BLEU & chrF++ & R-2 & R-L & MET. & P & R & F1 \\
\midrule
\multirow{6}{*}{Gemini 3 Flash} & Barisal & 44.3 & 71.42 & 40.4 & 67.3 & 65.3 & 88.7 & 85.2 & 86.5 \\
 & Chittagong & 46.2 & 71.26 & 42.6 & 63.1 & 63.4 & 62.7 & 59.8 & 58.2 \\
 & Noakhali & 41.5 & 73.46 & 47.4 & 68.8 & 67.6 & 73.1 & 65.9 & 65.9 \\
 & Rangpur & 25.4 & 62.00 & 41.9 & 63.6 & 62.1 & 80.2 & 70.4 & 70.3 \\
 & Sylhet & 38.4 & 67.30 & 39.6 & 60.5 & 60.0 & 72.8 & 66.6 & 64.6 \\
 & Standard & 42.1 & 36.81 & 46.2 & 65.1 & 66.0 & 72.1 & 63.2 & 62.1 \\
\midrule
\multirow{6}{*}{GPT-4o-mini} & Barisal & 40.7 & 65.75 & 39.5 & 61.3 & 59.3 & 76.8 & 78.7 & 77.2 \\
 & Chittagong & 29.1 & 59.26 & 35.1 & 55.0 & 54.6 & 60.1 & 60.0 & 59.3 \\
 & Noakhali & 29.6 & 67.90 & 43.3 & 62.8 & 62.8 & 71.0 & 71.6 & 69.3 \\
 & Rangpur & 26.6 & 60.18 & 33.9 & 54.3 & 53.5 & 65.7 & 65.9 & 65.8 \\
 & Sylhet & 26.9 & 58.58 & 35.3 & 53.3 & 52.6 & 65.7 & 64.7 & 63.9 \\
 & Standard & 35.1 & 36.18 & 39.1 & 57.1 & 58.0 & 62.1 & 57.1 & 57.5 \\
\midrule
\multirow{6}{*}{Claude Haiku 4.5} & Barisal & 44.1 & 70.10 & 42.2 & 65.5 & 63.1 & 84.8 & 78.2 & 80.0 \\
 & Chittagong & 32.3 & 64.19 & 33.8 & 58.8 & 58.5 & 63.3 & 62.6 & 62.4 \\
 & Noakhali & 29.7 & 63.91 & 34.8 & 57.2 & 57.0 & 69.4 & 64.3 & 64.2 \\
 & Rangpur & 28.3 & 60.73 & 37.8 & 59.7 & 58.7 & 77.5 & 71.0 & 71.2 \\
 & Sylhet & 40.7 & 70.32 & 43.8 & 63.4 & 63.0 & 72.5 & 69.8 & 69.2 \\
 & Standard & 35.5 & 37.30 & 38.5 & 57.0 & 57.5 & 67.1 & 59.3 & 58.2 \\
\midrule
\multirow{6}{*}{Qwen-3-4B} & Barisal & 6.4 & 32.28 & 9.2 & 20.7 & 27.0 & 17.8 & 50.0 & 26.3 \\
 & Chittagong & 8.9 & 33.48 & 7.4 & 22.5 & 24.7 & 67.3 & 54.5 & 42.5 \\
 & Noakhali & 6.1 & 34.73 & 6.7 & 22.1 & 21.7 & 58.3 & 53.6 & 41.0 \\
 & Rangpur & 6.3 & 36.57 & 13.5 & 26.0 & 30.3 & 71.5 & 50.9 & 31.8 \\
 & Sylhet & 0.0 & 37.34 & 6.6 & 21.1 & 22.8 & 75.8 & 55.8 & 44.4 \\
 & Standard & 10.5 & 23.49 & 15.2 & 32.1 & 33.0 & 52.2 & 49.1 & 47.3 \\
\midrule
\multirow{6}{*}{Gemma-4-4B} & Barisal & 40.7 & 70.71 & 39.3 & 63.0 & 62.7 & 84.9 & 84.9 & 84.9 \\
 & Chittagong & 32.4 & 62.56 & 33.6 & 58.4 & 57.9 & 64.8 & 64.7 & 64.4 \\
 & Noakhali & 29.8 & 63.22 & 32.6 & 55.6 & 57.3 & 68.9 & 68.1 & 68.4 \\
 & Rangpur & 15.0 & 49.63 & 26.5 & 46.7 & 47.5 & 71.4 & 69.0 & 69.2 \\
 & Sylhet & 27.8 & 60.65 & 32.9 & 52.2 & 53.3 & 76.6 & 76.0 & 76.0 \\
 & Standard & 32.2 & 35.03 & 37.1 & 55.2 & 55.5 & 66.5 & 62.2 & 63.1 \\
\midrule
\multirow{6}{*}{Llama-3.1-8B} & Barisal & 39.4 & 66.45 & 36.5 & 61.9 & 60.4 & 82.7 & 82.7 & 82.7 \\
 & Chittagong & 36.9 & 65.37 & 39.5 & 62.4 & 60.5 & 60.3 & 59.5 & 57.9 \\
 & Noakhali & 31.3 & 67.79 & 42.7 & 63.9 & 61.8 & 68.2 & 69.0 & 67.9 \\
 & Rangpur & 26.1 & 61.19 & 39.0 & 60.7 & 60.1 & 74.8 & 75.1 & 74.9 \\
 & Sylhet & 36.3 & 61.78 & 36.3 & 56.2 & 55.4 & 69.1 & 67.7 & 66.9 \\
 & Standard & 39.1 & 35.01 & 45.5 & 62.2 & 62.5 & 69.1 & 65.2 & 66.1 \\
\midrule
\multirow{6}{*}{Mistral-7B} & Barisal & 19.4 & 48.02 & 20.0 & 45.0 & 44.2 & 54.2 & 54.1 & 54.1 \\
 & Chittagong & 9.9 & 37.67 & 12.8 & 37.5 & 36.2 & 60.3 & 60.3 & 60.3 \\
 & Noakhali & 11.7 & 44.55 & 21.7 & 40.8 & 38.3 & 54.8 & 55.0 & 54.1 \\
 & Rangpur & 9.2 & 43.21 & 19.3 & 40.2 & 41.1 & 55.1 & 55.2 & 55.0 \\
 & Sylhet & 10.8 & 44.59 & 12.7 & 34.1 & 35.8 & 64.5 & 64.4 & 64.4 \\
 & Standard & 12.2 & 26.12 & 17.1 & 32.2 & 33.1 & 57.1 & 55.2 & 55.5 \\
\bottomrule
\end{tabular}
\caption{Chain-of-thought transliteration task results (Romanized output from native-script input). The same model run produces both this transliteration output and the English translation in Table~\ref{tab:app_cot_native}; the two share the subjectivity (P, R, F1) classifier output. chrF++ is the appropriate character-level metric for a Romanization target; COMET is not applicable here, as its estimator is trained for translation into a natural language rather than for a fixed transliteration convention. The Standard Bangla row is the low chrF++ outlier for every model because transliterating Standard Bangla carries no dialectal phonology to preserve, so models default to a conventional Romanization that diverges from the Avro reference convention---a property of the reference scheme, not a model failure.}
\label{tab:app_cot_translit}
\end{table*}

\begin{table*}[!tp]
\centering
\small
\begin{tabular}{llccccccc}
\toprule
Model & Dialect & BLEU & R-2 & R-L & MET. & P & R & F1 \\
\midrule
\multirow{6}{*}{Qwen-3-4B} & Barisal & 84.7 & 86.6 & 92.9 & 93.3 & 85.7 & 91.0 & 87.8 \\
 & Chittagong & 45.5 & 47.9 & 62.7 & 64.0 & 69.7 & 70.0 & 69.7 \\
 & Noakhali & 52.7 & 55.0 & 71.1 & 70.8 & 71.8 & 71.8 & 71.8 \\
 & Rangpur & 65.7 & 67.6 & 78.4 & 79.2 & 78.3 & 78.3 & 78.0 \\
 & Sylhet & 59.0 & 65.5 & 76.7 & 77.6 & 80.4 & 79.4 & 79.6 \\
 & Standard & 87.8 & 88.8 & 94.8 & 93.4 & 77.5 & 76.5 & 76.9 \\
\midrule
\multirow{6}{*}{Gemma-4-4B} & Barisal & 86.7 & 87.3 & 93.4 & 94.0 & 74.7 & 83.9 & 76.2 \\
 & Chittagong & 41.9 & 44.9 & 60.6 & 59.7 & 61.3 & 61.5 & 60.9 \\
 & Noakhali & 52.4 & 56.8 & 70.4 & 70.3 & 71.8 & 71.8 & 71.8 \\
 & Rangpur & 54.0 & 59.6 & 72.6 & 73.3 & 78.3 & 78.3 & 78.0 \\
 & Sylhet & 57.2 & 63.5 & 75.5 & 76.0 & 78.2 & 78.3 & 78.0 \\
 & Standard & 91.2 & 92.6 & 97.2 & 96.4 & 80.0 & 78.2 & 78.8 \\
\midrule
\multirow{6}{*}{Llama-3.1-8B} & Barisal & 85.9 & 87.0 & 92.5 & 93.0 & 76.2 & 85.2 & 78.1 \\
 & Chittagong & 52.2 & 52.8 & 66.1 & 66.2 & 71.0 & 71.4 & 70.9 \\
 & Noakhali & 58.9 & 63.7 & 76.7 & 75.9 & 77.3 & 78.9 & 78.1 \\
 & Rangpur & 66.9 & 68.8 & 79.2 & 79.6 & 81.2 & 81.2 & 81.0 \\
 & Sylhet & 59.3 & 62.0 & 76.2 & 75.8 & 80.0 & 79.7 & 79.8 \\
 & Standard & 90.4 & 91.7 & 97.0 & 95.6 & 81.8 & 80.7 & 81.1 \\
\midrule
\multirow{6}{*}{Mistral-7B} & Barisal & 85.6 & 87.7 & 93.7 & 94.1 & 74.0 & 83.3 & 75.2 \\
 & Chittagong & 61.0 & 62.7 & 72.4 & 72.5 & 67.3 & 67.6 & 66.9 \\
 & Noakhali & 66.8 & 67.0 & 79.3 & 78.8 & 77.4 & 80.5 & 78.7 \\
 & Rangpur & 71.9 & 74.5 & 85.1 & 84.6 & 80.6 & 80.4 & 80.0 \\
 & Sylhet & 63.6 & 68.5 & 80.0 & 79.0 & 87.0 & 86.8 & 86.9 \\
 & Standard & 92.9 & 94.1 & 97.7 & 96.2 & 83.6 & 83.1 & 83.3 \\
\bottomrule
\end{tabular}
\caption{LoRA fine-tuning per-dialect results, native Bangla script, open-source models only. Hyperparameters: rank $r=16$, $\alpha=32$, dropout $0.05$, AdamW learning rate $2\!\times\!10^{-4}$, $3$ epochs, batch $4$, gradient accumulation $4$ (effective batch $16$); $5\%$ warmup.}
\label{tab:app_lora_native}
\end{table*}

\begin{table*}[!tp]
\centering
\small
\begin{tabular}{llccccccc}
\toprule
Model & Dialect & BLEU & R-2 & R-L & MET. & P & R & F1 \\
\midrule
\multirow{6}{*}{Qwen-3-4B} & Barisal & 72.7 & 74.6 & 80.9 & 81.3 & 77.7 & 83.0 & 79.8 \\
 & Chittagong & 33.5 & 35.9 & 50.7 & 52.0 & 61.7 & 62.0 & 61.7 \\
 & Noakhali & 40.7 & 43.0 & 59.1 & 58.8 & 63.8 & 63.8 & 63.8 \\
 & Rangpur & 53.7 & 55.6 & 66.4 & 67.2 & 70.3 & 70.3 & 70.0 \\
 & Sylhet & 47.0 & 53.5 & 64.7 & 65.6 & 72.4 & 71.4 & 71.6 \\
 & Standard & 75.8 & 76.8 & 82.8 & 81.4 & 69.5 & 68.5 & 68.9 \\
\midrule
\multirow{6}{*}{Gemma-4-4B} & Barisal & 74.7 & 75.3 & 81.4 & 82.0 & 66.7 & 75.9 & 68.2 \\
 & Chittagong & 29.9 & 32.9 & 48.6 & 47.7 & 53.3 & 53.5 & 52.9 \\
 & Noakhali & 40.4 & 44.8 & 58.4 & 58.3 & 63.8 & 63.8 & 63.8 \\
 & Rangpur & 42.0 & 47.6 & 60.6 & 61.3 & 70.3 & 70.3 & 70.0 \\
 & Sylhet & 45.2 & 51.5 & 63.5 & 64.0 & 70.2 & 70.3 & 70.0 \\
 & Standard & 79.2 & 80.6 & 85.2 & 84.4 & 72.0 & 70.2 & 70.8 \\
\midrule
\multirow{6}{*}{Llama-3.1-8B} & Barisal & 73.9 & 75.0 & 80.5 & 81.0 & 68.2 & 77.2 & 70.1 \\
 & Chittagong & 40.2 & 40.8 & 54.1 & 54.2 & 63.0 & 63.4 & 62.9 \\
 & Noakhali & 46.9 & 51.7 & 64.7 & 63.9 & 69.3 & 70.9 & 70.1 \\
 & Rangpur & 54.9 & 56.8 & 67.2 & 67.6 & 73.2 & 73.2 & 73.0 \\
 & Sylhet & 47.3 & 50.0 & 64.2 & 63.8 & 72.0 & 71.7 & 71.8 \\
 & Standard & 78.4 & 79.7 & 85.0 & 83.6 & 73.8 & 72.7 & 73.1 \\
\midrule
\multirow{6}{*}{Mistral-7B} & Barisal & 73.6 & 75.7 & 81.7 & 82.1 & 66.0 & 75.3 & 67.2 \\
 & Chittagong & 49.0 & 50.7 & 60.4 & 60.5 & 59.3 & 59.6 & 58.9 \\
 & Noakhali & 54.8 & 55.0 & 67.3 & 66.8 & 69.4 & 72.5 & 70.7 \\
 & Rangpur & 59.9 & 62.5 & 73.1 & 72.6 & 72.6 & 72.4 & 72.0 \\
 & Sylhet & 51.6 & 56.5 & 68.0 & 67.0 & 79.0 & 78.8 & 78.9 \\
 & Standard & 80.9 & 82.1 & 85.7 & 84.2 & 75.6 & 75.1 & 75.3 \\
\bottomrule
\end{tabular}
\caption{LoRA fine-tuning per-dialect results, transliterated input. Adapters are trained from scratch on the Romanized fold of the LoRA training set ($160$ training instances per dialect).}
\label{tab:app_lora_translit}
\end{table*}

\end{document}